%% file: root.tex
\documentclass[letterpaper, 10 pt, conference]{ieeeconf}  

\IEEEoverridecommandlockouts                              

\usepackage{graphicx}
\usepackage{float}
\usepackage{balance}
\usepackage[hidelinks]{hyperref}
\hypersetup{
  pdftitle={MoMa-NMP: Scene-Scale Neural Motion Planning for Mobile Manipulators},
  pdfauthor={Adam Kan, Jason Jingzhou Liu, Jiahui Yang, Kenneth Shaw, Deepak Pathak}
}
\usepackage{amsmath} 
\usepackage{amssymb}  

\title{\LARGE \bf
MoMa-NMP: Scene-Scale Neural Motion Planning for Mobile Manipulators
}

\author{Adam Kan, Jason Jingzhou Liu, Jiahui Yang, Kenneth Shaw, and Deepak Pathak%
\thanks{All authors are with Carnegie Mellon University, Pittsburgh, PA, USA. Corresponding author: Adam Kan (\texttt{ahkan@andrew.cmu.edu}).}%
}

\begin{document}

\maketitle
\thispagestyle{empty}
\pagestyle{empty}

\begin{abstract}
Mobile manipulators operating in homes and other complex environments must coordinate their base and arm over long horizons, navigating around obstacles, through narrow passages, and between constrained workspaces. Neural motion planners can provide fast closed-loop planning from partial observations, but scaling them to room-sized environments is difficult because generating high-quality training trajectories is itself a challenging motion-planning problem. To address this, we introduce MoMa-DemoSmith, a demonstration-generation pipeline that separates the problem into global path discovery for large complex scenes and local motion refinement for collision avoidance. This key separation enables the generation of millions of challenging whole-body trajectories across our procedurally generated scenes. We train MoMa-NMP on these demonstrations through behavior cloning (BC) followed by privileged on-policy distillation, achieving 87.0\% success on our simulation benchmark and 77.5\% across 80 trials in unseen real-world scenes. See video results at \href{https://adamhkan.com/moma-nmp/}{\nolinkurl{adamhkan.com/moma-nmp/}}.
\end{abstract}

\section{Introduction}

\begin{figure*}[!t]
\centering
\includegraphics[width=\textwidth]{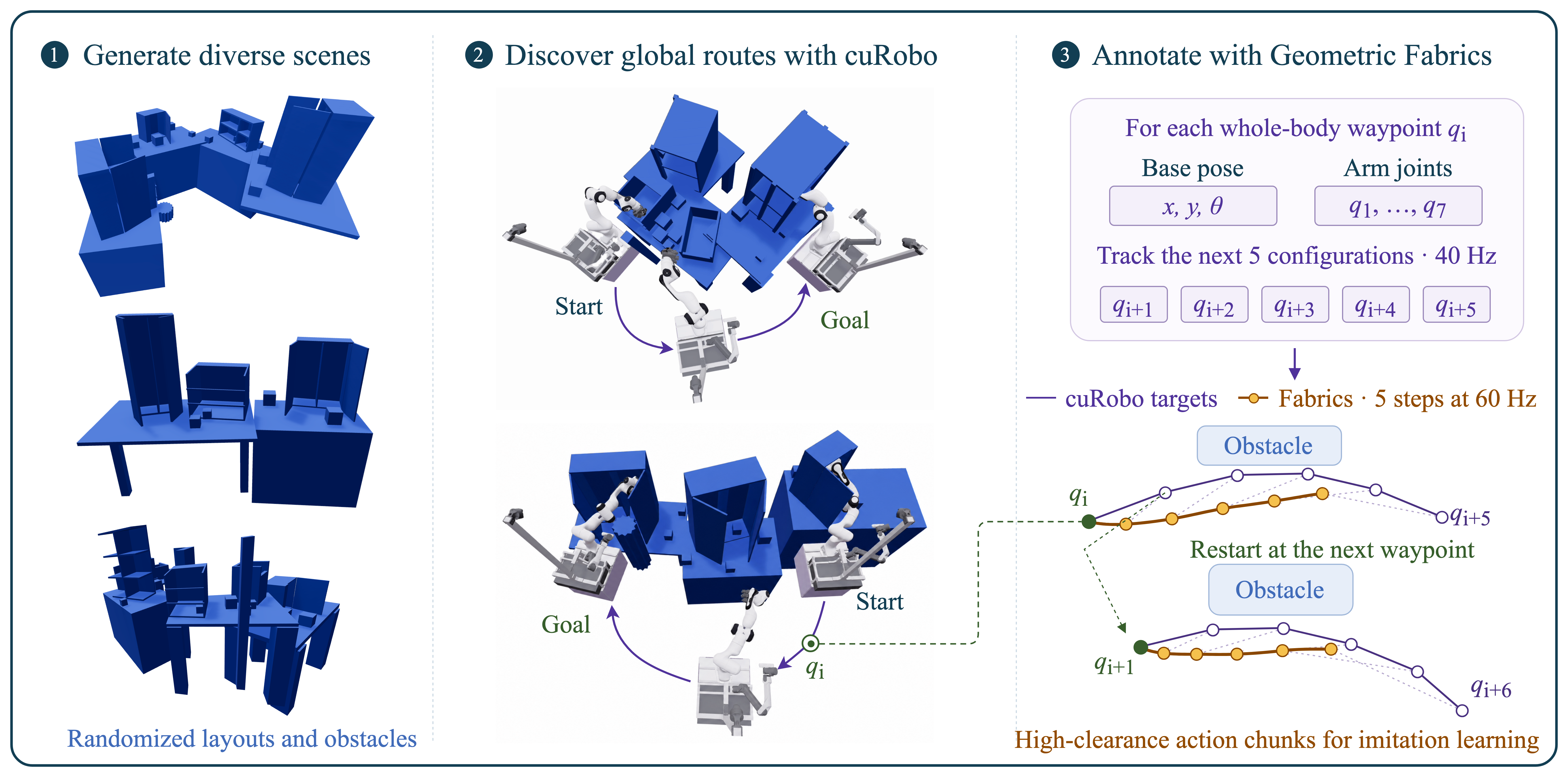}
\caption{MoMa-DemoSmith demonstration generation. (1) We generate diverse scenes with varying arrangements of tables and obstacles, and sample start and goal poses in both open and confined workspaces. (2) We use cuRobo to plan global trajectories for the mobile base and arm. Relaxed planning constraints improve route discovery but can produce trajectories with insufficient obstacle clearance or unrealistic base motion. (3) We use Geometric Fabrics to refine the next five whole-body configuration targets at each cuRobo waypoint, generating action chunks with improved obstacle clearance for imitation learning.}
\label{fig:demosmith}
\end{figure*}

Robots operating throughout homes, offices, and other complex environments must coordinate mobility and manipulation across entire scenes. They must withdraw an end-effector from confined spaces, navigate around obstacles, coordinate whole-body motion through narrow passages, and reposition to access new workspaces. These capabilities require precise whole-body control, long-horizon planning, and global reasoning under partial observability.

Classical motion planners have long been used during robot deployment to address such problems. However, classical planners generally rely on explicit scene geometry and can become computationally expensive as scene scale, planning horizon, and robot dimensionality increase. Neural motion planners offer an alternative: they can generate actions with low latency, operate directly from partial observations, and replan continuously at runtime as new geometry becomes visible.

Training such neural motion planners, however, requires large datasets that are particularly difficult to generate at room scale. Demonstrations must capture long, nonlinear whole-body motions around multiple obstacles while remaining collision-free and suitable for imitation. Classical motion planners can provide this supervision, but generating high-quality trajectories reliably and at sufficient throughput becomes increasingly difficult as scene scale and geometric constraints grow.

To address this challenge, we introduce MoMa-DemoSmith, a parallelizable demonstration-generation pipeline for room-scale mobile manipulation. DemoSmith first procedurally constructs multi-workspace environments with dense obstacles and constrained start and goal regions. It must then generate learning-ready whole-body trajectories through these environments. However, cuRobo \cite{sundaralingam2023curobo}, a state-of-the-art GPU-accelerated motion-generation framework, often struggles to discover the long, nonlinear base trajectories required to maneuver around obstacles in complex scenes.

Our key insight is to relax cuRobo’s collision margin and velocity limits to improve global path discovery, allowing it to identify the high-level whole-body route between start and goal configurations. However, these relaxed constraints can produce trajectories that pass too close to obstacles and contain unrealistic base velocities. We therefore use Geometric Fabrics \cite{vanwyk2022fabrics}, a reactive motion generator with access to ground-truth obstacle geometry, to generate high-clearance action targets along the cuRobo route for imitation learning.

Finally, we use these demonstrations to train a closed-loop neural policy, MoMa-NMP. We first use behavior cloning to train a policy from demonstrations generated by MoMa-DemoSmith. However, over long, scene-spanning trajectories, small prediction errors can compound and drive the policy into collisions. During on-policy fine-tuning, we use Geometric Fabrics to correct action chunks predicted by a frozen BC policy. Through on-policy student-teacher distillation, these corrected actions supervise a student policy that operates on partially observable scenes in deployment.

To our knowledge, MoMa-NMP is the first general-purpose neural motion planner to demonstrate closed-loop whole-body planning for a mobile manipulator in unseen cluttered real-world scenes. The policy learns to execute globally constrained motions---coordinating base and arm around multiple obstacles and into narrow or initially occluded workspaces---using only partial onboard observations, without a precomputed global route or task-specific training. Across 80 trials spanning 20 tasks and 10 unseen scenes, MoMa-NMP achieves 77.5\% success.

\section{Related Work}
\textbf{Classical and reactive motion planning.}
Robot motion planning has traditionally relied on geometric search and trajectory optimization. Sampling-based methods such as PRM and RRT-Connect provide global reasoning over complex environments \cite{kavraki1996prm,kuffner2000rrtconnect}, while CHOMP, STOMP, TrajOpt, and GPMP2 optimize collision-free trajectories \cite{ratliff2009chomp,kalakrishnan2011stomp,schulman2014trajopt,mukadam2018gpmp2}. GPU-accelerated systems such as cuRobo further improve planning throughput \cite{sundaralingam2023curobo}. However, planning becomes more expensive as robot dimensionality and scene complexity increase. Reactive methods such as RMPflow and Geometric Fabrics generate local motion at high control rates by combining goal-attraction, obstacle-avoidance, and other task-space objectives into smooth joint-space acceleration policies \cite{cheng2021rmpflow,vanwyk2022fabrics}. However, because they do not perform global search, they can become trapped in local minima when reaching the goal requires a long, non-local route.

\textbf{Learning-augmented motion planning.}
A complementary line of work uses learned models to accelerate classical planning rather than replacing it. Learned samplers bias sampling-based planners toward promising regions of configuration space \cite{ichter2018sampling,qureshi2018deepsmp,lai2021plannerflows,johnson2023vqmpt}, while other methods learn critical regions or states that are explicitly incorporated into sampling-based search \cite{molina2020learnlink,ichter2020criticalprm}. These approaches retain explicit global search while using learning to reduce the portion of configuration space that must be explored. For whole-body planning, Cheng et al.\ learn a sampling distribution for a high-DoF mobile robot while retaining classical graph search \cite{cheng2020sampling}, and Neural Randomized Planning (NRP) combines a global sampling-based planner with local neural samplers whose distributions are composed through the evolving search structure \cite{lu2024nrp}. Such hybrid methods preserve the global reasoning of classical planning, but still perform explicit search for each new planning problem.

\textbf{Neural motion planning from scene observations.} Rather than using learning to guide an online search procedure, neural motion planners can amortize the planning computation itself into a policy. MPNet recursively predicts intermediate configurations from environment observations and can also be combined with classical sampling-based planners for additional guarantees \cite{qureshi2019mpnet}. M$\pi$Nets demonstrated closed-loop planning for fixed-base manipulators from partial depth observations \cite{fishman2023mpinets}, while diffusion-based methods model multimodal trajectory distributions \cite{yan2025m2diffuser,huang2023scenediffuser,carvalho2023motiondiffusion}. Neural MP scales neural planning with large procedurally generated datasets \cite{dalal2025neuralmp}, while Deep Reactive Policy improves closed-loop robustness through student-teacher fine-tuning on states encountered during execution \cite{yang2025drp}.

\textbf{Whole-body and scene-scale planning.}
Recent work extends planning and control from fixed-base arms to mobile manipulators. Coupled base-arm planning and MotionBenchMaker provide foundations for mobile planning and large-scale dataset generation \cite{jiao2021vkc,chamzas2022motionbenchmaker}. Spahn et al.\ perform coupled base-arm trajectory optimization with free-space decomposition \cite{d901a4cef1c6447c99e28ced87264572}, while Zheng et al.\ develop reactive whole-body control with link-specific collision constraints \cite{994917fa14d74d25b22c7a0364270210}. Among learned approaches, N$^2$M$^2$ learns mobile-manipulator navigation conditioned on externally generated EEF motions \cite{honerkamp2022learning}. ActPerMoMa couples mobile motion with active perception for manipulation in cluttered scenes \cite{jauhri2024actpermoma}, while Mobi-$\pi$ selects mobile-base poses that enable pretrained manipulation policies to operate from favorable viewpoints \cite{yang2025mobipi}. HoMeR learns long-range mobile manipulation through hybrid imitation learning and whole-body control \cite{sundaresan2026homer}. M$^3$Bench introduces large-scale whole-body tasks and demonstrations \cite{zhang2025m3bench}, and AutoMoMa scales whole-body trajectory generation using augmented kinematic representations and GPU-parallel optimization \cite{niu2026scalable}. M$^2$Diffuser learns coordinated base-and-arm trajectories from 3D scene observations \cite{yan2025m2diffuser}.

\section{Method}\label{sec:method}
We consider a mobile manipulator consisting of a modified TidyBot++ \cite{wu2024tidybot++} holonomic mobile base, a Franka Emika Panda robot arm, and a LEAP Hand \cite{shaw2023leap} held in a fixed configuration. The robot is equipped with a RealSense depth camera and a LiDAR, which provide the partial scene observations used by MoMa-NMP.

We formulate whole-body motion planning as a goal-conditioned policy that maps the current robot state and a partial scene observation to a sequence of base-and-arm actions. At time $t$, the policy observes the seven Franka joint angles $\mathbf{q}^{\mathrm{arm}}_t$, a point-cloud observation $P_t$, the current end-effector (EEF) pose $C_t$, and a goal EEF pose $G_t$. All geometric quantities are expressed relative to the robot base at the time of inference. The policy receives no world-frame base pose. The policy predicts a five-step action chunk
\begin{equation}
A_t =
[\Delta \mathbf{q}^{1}_t,\ldots,\Delta \mathbf{q}^{H}_t],
\qquad H=5,
\end{equation}
where each $\Delta \mathbf{q}^{i}_t \in \mathbb{R}^{10}$ specifies offsets for base $x$, $y$, and yaw and the seven Franka joints relative to the configuration observed at time $t$. Base offsets are expressed in the current robot-base frame.

Our method addresses two challenges in learning scene-scale motion planning. MoMa-DemoSmith first generates learnable demonstrations for planning problems that are difficult even for classical experts. We then train MoMa-NMP by behavior cloning these demonstrations and use privileged on-policy fine-tuning to improve closed-loop execution from partial observations.

\subsection{MoMa-DemoSmith}

MoMa-DemoSmith generates demonstrations for scene-scale planning across large and diverse environments as seen in Figure \ref{fig:demosmith}. It procedurally constructs multi-workspace scenes with multiple tables, obstacles, and constrained start and goal regions, producing tasks that require long, nonlinear whole-body motions between distinct workspaces. These problems are difficult for a single classical planner to both solve reliably and convert into high-quality supervision. DemoSmith therefore separates trajectory generation into two stages: cuRobo discovers a globally feasible route through the scene, and Geometric Fabrics refines that route into high-clearance action targets for imitation learning.

\subsubsection{Procedural Scene and Task Generation}

To generate a wide range of room-scale planning problems, DemoSmith constructs each environment using one to seven tables. Tables are populated with randomized assets and obstacles, then arranged in a line, inward-facing circle, or outward-facing circle with approximately equal probability. An additional pillar is placed near each table independently with probability $0.33$. These layouts span open environments and tightly constrained scenes in which surrounding geometry strongly determines the mobile base's global route.

To create motions that begin and end in constrained workspaces, we sample start and goal EEF poses both in free space and inside compartments associated with scene assets. We define compartments as enclosed manipulation regions such as the space inside drawers, microwaves, cabinets, or shelves. Compartment-to-compartment problems can therefore require the robot to withdraw from one tight workspace, traverse the scene, and enter another. We use cuRobo \cite{sundaralingam2023curobo}, a GPU-accelerated motion-generation framework that combines parallel geometric planning and trajectory optimization and also provides collision-aware inverse kinematics (IK), to solve IK for each sampled pose.

For the goal configuration, we additionally prefer solutions that keep the Franka away from joint limits and collisions while orienting the base toward the relevant workspace, which keeps the onboard depth camera directed toward nearby scene geometry. Starting from the cuRobo IK solution, we apply a Riemannian Motion Policy (RMP)-based optimizer \cite{ratliff2018riemannian}, a framework for combining attractor and avoidance objectives into a single robot motion, to preserve the target EEF pose while attracting the Franka toward a nominal retract posture and orienting the base toward the nearest table. Because this stage does not guarantee a collision-free configuration, we perform a final cuRobo IK solve initialized from the regularized solution.

\subsubsection{Global Path Discovery and Fabrics Action Annotation}

DemoSmith scenes often require long, nonlinear base motions around multiple obstacles. Under realistic collision and motion limits, cuRobo frequently fails to discover these constrained whole-body routes. Relaxing these limits improves global path discovery, but the resulting trajectories can pass too close to obstacles and exhibit unrealistic base velocities, making them poor imitation targets.

DemoSmith therefore separates global path discovery from local motion refinement. cuRobo is used to identify a feasible global route through the scene, while Geometric Fabrics follows that route to produce higher-clearance, more realistic action targets for imitation learning.

For the cuRobo global-planning stage, we therefore use a $4\,\mathrm{mm}$
collision buffer and set the holonomic base $x$, $y$, and yaw velocity limits
to $6\times$ the corresponding hardware limits, while keeping the Franka
limits near their real values. Base acceleration and jerk limits are also
relaxed. We select cuRobo based on its performance in our six-planner backend benchmark, where it achieves $1.86\times$ the throughput of RRTConnect, the next-fastest backend under calibrated parallelism (Appendix~\ref{app:planner_backend}).

To convert the cuRobo paths into lower-magnitude action targets that maintain higher clearance from obstacles, we use Geometric Fabrics \cite{vanwyk2022fabrics}, an acceleration-based reactive motion-generation framework that composes objectives such as target attraction, obstacle avoidance, and joint-limit avoidance. cuRobo trajectories are sampled at $40\,\mathrm{Hz}$. At each cuRobo waypoint, we initialize Fabrics at the corresponding whole-body configuration and provide the next five cuRobo configurations sequentially as joint-space targets. We roll out Fabrics for five steps at $60\,\mathrm{Hz}$, producing the five-step action chunk used as the imitation target. This differs from the Hybrid Planner in M$\pi$Nets \cite{fishman2023mpinets}, which provides Geometric Fabrics with only EEF waypoints from the global planner. These EEF-only targets do not constrain intermediate arm configurations; in our mobile-manipulator setting, they would additionally leave the base pose unconstrained. In DemoSmith's long-horizon, tightly constrained scenes, this can pull the robot away from the globally feasible route and into a local minimum. We therefore provide the next five full joint-space configurations from cuRobo, including both the base pose and Franka joints, as successive Fabrics targets. This preserves cuRobo's global path while allowing Fabrics to improve local clearance and motion quality.

Using this pipeline, we generate approximately 6.7 million global trajectories with local Fabrics action annotations. Global-trajectory generation produces approximately 465 accepted trajectories per GPU-hour, and Fabrics annotation processes approximately 36,000 precomputed records per GPU-hour on an RTX A5000 (Appendix~\ref{app:demosmith_generation_throughput}).

\begin{figure}[H]
\centering
\includegraphics[width=\columnwidth]{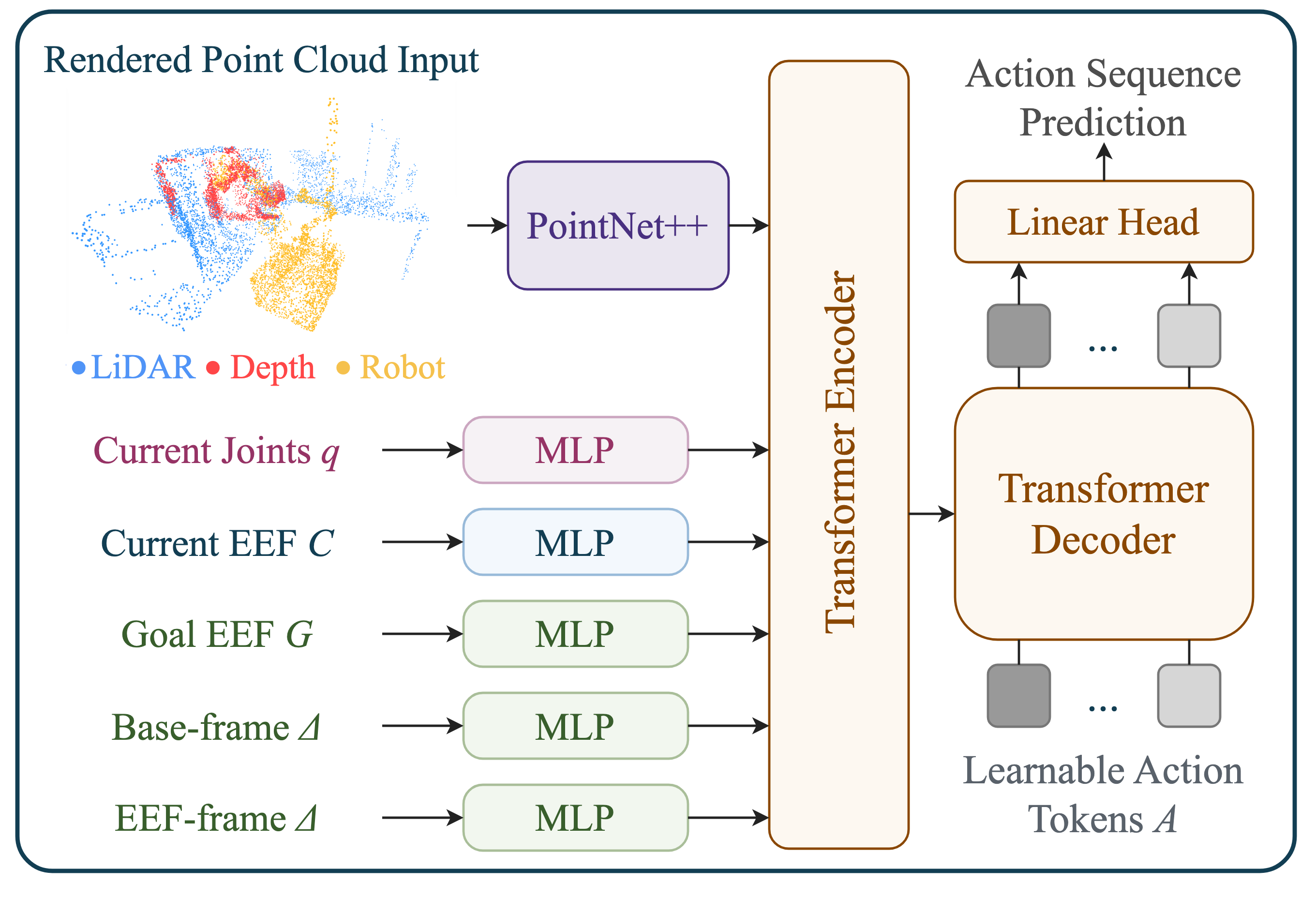}
\caption{MoMa-NMP combines point-cloud representations with the current robot state and goal-pose representations into a transformer encoder. A decoder with learnable action tokens predicts a five-step chunk of joint-space actions.}
\label{fig:model_architecture}
\end{figure}

\subsection{MoMa-NMP}

MoMa-DemoSmith provides high-quality demonstrations of difficult scene-scale planning problems, making behavior cloning a natural first step for training MoMa-NMP. However, over long scene-spanning trajectories, small prediction errors compound and push the policy into states not represented in the offline demonstration distribution. We therefore train MoMa-NMP in two stages: we first train the policy using behavior cloning on DemoSmith demonstrations, then apply privileged on-policy fine-tuning on states encountered during closed-loop rollouts.

In the second stage, we use Geometric Fabrics in a second, distinct role. In DemoSmith, Fabrics converts cuRobo paths into offline imitation targets; during fine-tuning, Fabrics instead acts as a privileged teacher that locally corrects actions proposed by a frozen behavior-cloned base policy using ground-truth scene geometry. Only the neural policy and onboard observations are used at deployment.
\subsubsection{Policy Architecture and Goal Representation}

MoMa-NMP uses a transformer-based action-chunk architecture adapted from IMPACT \cite{yang2025drp} as seen in Figure \ref{fig:model_architecture}. The point-cloud observation is encoded with PointNet++ \cite{qi2017pointnet++}, while the Franka joint configuration and goal-related quantities are independently embedded with MLPs. These features are processed by a transformer encoder-decoder, and a shared linear head predicts a five-step whole-body action chunk.

In large, partially observed scenes, a motion planner cannot generally assume access to a collision-free goal robot configuration. Instead, MoMa-NMP is conditioned on an EEF goal. To provide complementary representations of the transform between the current and goal EEF poses, we augment the current EEF pose $C$ and goal pose $G$ with left- and right-multiplicative SE(3) pose differences:
\begin{equation}
\Delta_{\mathrm{base}} = GC^{-1},
\qquad
\Delta_{\mathrm{eef}} = C^{-1}G.
\end{equation}
Both $C$ and $G$ are expressed in the current robot-base frame. Thus, $\Delta_{\mathrm{base}}$ is the left-multiplicative pose difference between the current and goal EEF poses in their shared robot-base coordinate convention, while $\Delta_{\mathrm{eef}}$ expresses the goal pose relative to the current EEF frame. Each SE(3) transform is encoded as a 9D vector containing the first two columns of its rotation matrix followed by its translation. Because all of these quantities are recomputed relative to the current base, the policy never observes an absolute base position or heading.

\subsubsection{Partial Point-Cloud Observations and Behavior Cloning}

At deployment, MoMa-NMP plans from onboard depth-camera and LiDAR observations rather than privileged scene geometry. We reproduce this partial observability during training with GPU-based sensor rendering. We sample scene and robot geometry, then project the combined points through depth-camera and LiDAR renderers that retain only the nearest return in each image pixel or angular bin. Occluded geometry is therefore removed before the policy observes the scene. Depth-camera and LiDAR returns are combined into a single partial-scene point cloud. A subsample of the known robot geometry is then appended, and a binary per-point feature distinguishes partial-scene points from robot points before PointNet++ encoding.

We first train MoMa-NMP with behavior cloning on approximately 6.7 million DemoSmith trajectories. Training examples consist of Fabrics-annotated five-step action chunks sampled from the demonstrations, and the policy is optimized with an $L_2$ loss over valid entries in the five predicted 10-DoF offsets. Because the most constrained motion often occurs while exiting the start region or approaching the goal, we sample the first and last 20 timestamps of each trajectory with twice the probability of intermediate timestamps.

\subsubsection{Privileged On-Policy Fine-Tuning}

Behavior cloning alone is insufficient for the long-horizon, tightly constrained trajectories considered in this work. Small prediction errors compound during closed-loop execution and push the policy into states not represented in the offline demonstration distribution. We therefore add a second, on-policy training stage that provides corrective supervision on states encountered during policy rollouts. We build on DRP's privileged student-teacher formulation, using Geometric Fabrics as the local corrective teacher while the frozen BC policy supplies the global motion intent. Fabrics uses privileged scene geometry to supply these corrections.

Both the trainable MoMa-NMP student and a frozen base policy are initialized from the same behavior-cloned checkpoint. At each visited state, the base policy predicts a full five-step action chunk. Unlike DRP, which refines only the first predicted joint target, we provide all five whole-body targets sequentially to Geometric Fabrics and use the resulting five-step corrected chunk to supervise the student. Once the EEF is close to the commanded goal, the remaining motion is a local planning problem that Fabrics can solve directly. We use the same goal-locking strategy as DRP: within $0.10\,\mathrm{m}$ and $30^\circ$ of the goal, the teacher activates EEF attraction while retaining the base policy's configuration targets as posture guidance.

To progressively expose the student to its own state distribution, we begin with teacher-controlled rollouts and linearly transition toward student control after an initial warm-up period \cite{ross2010efficient}. Because MoMa-NMP is trained to avoid contact rather than model contact-rich dynamics, we perform these rollouts in a lightweight custom kinematic simulator rather than a full rigid-body simulator. The environment advances at $60\,\mathrm{Hz}$ by applying commanded configurations and checking collisions after each step. We run approximately 16k environments in parallel across eight RTX A6000 GPUs.

\section{Experiments and Results}

\begin{figure*}[!t]
\centering
\includegraphics[width=\textwidth]{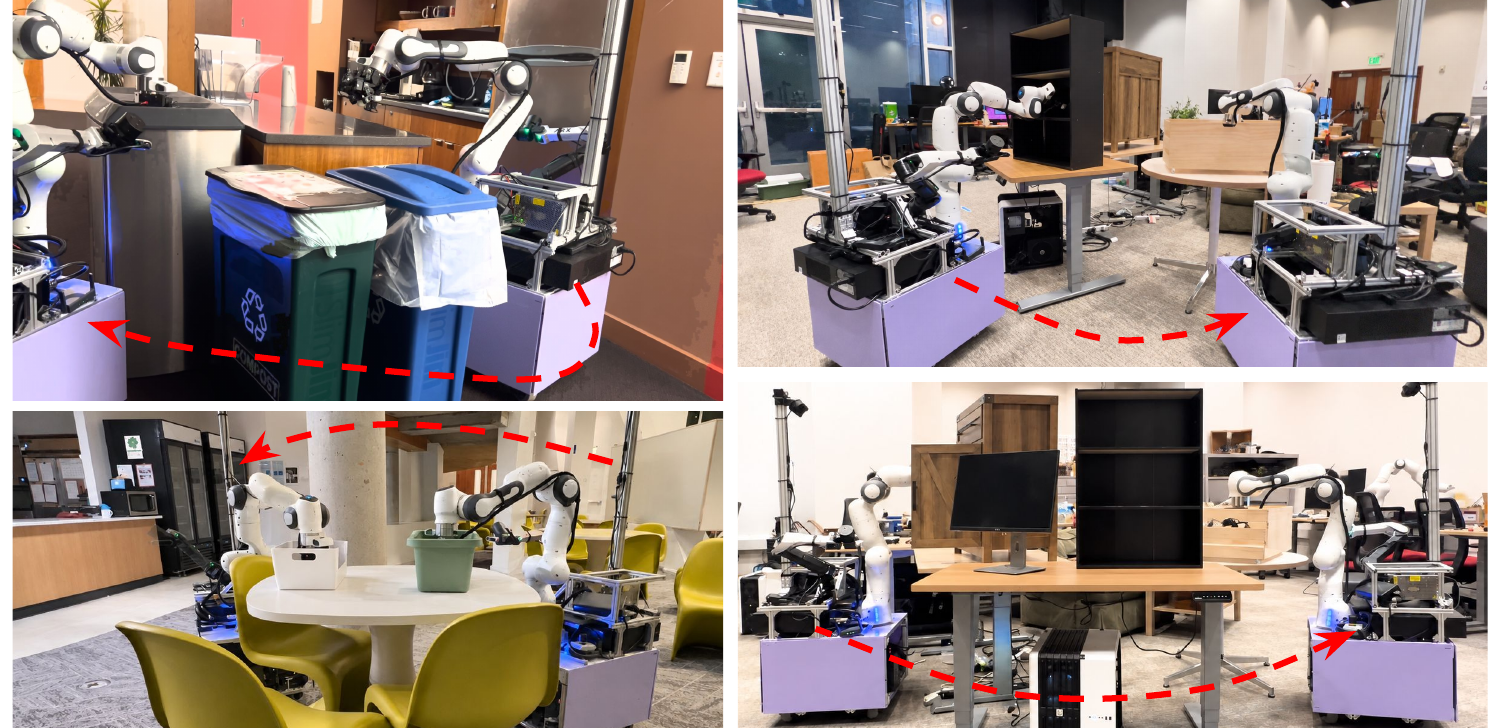}
\caption{MoMa-NMP plans collision-free trajectories between compartments in large scenes, navigating through tight spaces with nonlinear base motion.}

\label{fig:results}
\end{figure*}

\begin{table*}[!t]
\caption{Simulation Planning Results}
\label{tab:mobile_mp_simulation}
\centering
\small
\setlength{\tabcolsep}{5pt}
\begin{tabular}{lcccc}
\hline
Method &
\shortstack{Success\\(\%)} &
\shortstack{Comp.\ $\to$ comp.\\success (\%)} &
\shortstack{$\geq 4$ tables\\success (\%)} &
\shortstack{World collision\\(\%)} \\
\hline
MoMa-NMP (BC only) & 64.8 & 56.6 & 60.4 & 34.1 \\
MoMa-NMP & \textbf{87.0} & \textbf{84.4} & \textbf{85.1} & 12.3 \\
\hline
\multicolumn{5}{l}{\textit{With privileged scene and goal information:}} \\
Geometric Fabrics & 17.2 & 10.4 & 10.4 & 81.9 \\
cuRobo & 46.9 & 45.6 & 48.3 & \textbf{0.0} \\
OMPL RRTConnect & 85.3 & 83.9 & 84.8 & \textbf{0.0} \\
\hline
\end{tabular}
\par\smallskip
\begin{minipage}{\linewidth}
\footnotesize
MoMa-NMP plans from partial sensor observations. cuRobo and RRTConnect
receive oracle geometry, eight precomputed goal configurations,
and relaxed mobile-base motion limits, with 3~s and 60~s planning budgets,
respectively. Geometric Fabrics receives oracle geometry and runs reactively.
World collision denotes robot--environment collision before successful goal reach.
Robot link--link self-collision is not scored in this benchmark.
\end{minipage}

\end{table*}

We evaluate MoMa-NMP to answer three questions: (1) Can closed-loop neural motion planning solve difficult room-scale whole-body planning problems that challenge existing planning approaches? (2) Which data-generation and learning choices enable this capability? (3) Does the learned planner generalize to unseen real-world environments from partial onboard observations? Qualitative results are shown in Figure \ref{fig:results}, with additional videos available at \href{https://adamhkan.com/moma-nmp/}{\nolinkurl{adamhkan.com/moma-nmp/}}.

\subsection{Evaluation Setup}

\textbf{Simulation.}
We evaluate on 9,937 valid held-out start--goal problems drawn from successfully generated MoMa-DemoSmith demonstrations (Appendix~\ref{app:simulation_details}). A rollout succeeds when the EEF reaches within $1\,\mathrm{cm}$ and $15^{\circ}$ of the goal without a prior robot--environment collision. Robot link--link self-collision is not scored in this benchmark. In addition to aggregate success, we report compartment-to-compartment tasks and scenes containing at least four tables to isolate tightly constrained and larger planning problems.

\textbf{Real-world.}
We deploy MoMa-NMP on the mobile manipulator described in Section~\ref{sec:method}. The policy receives onboard RealSense depth-camera and LiDAR observations and replans at $12\,\mathrm{Hz}$. The policy commands the holonomic base and seven Franka joints, while the LEAP Hand remains fixed throughout the experiments. The five-step action chunk is tracked successively at $60\,\mathrm{Hz}$ without interpolation. EEF goals are selected in a 3D visualization of the observed scene point cloud. SLAM and wheel odometry track the base pose to keep the goals fixed in space and measure goal-reaching accuracy. We evaluate in 10 previously unseen real-world scenes containing geometries and arrangements not present in the procedurally generated training environments.

\textbf{Baselines.}
We compare MoMa-NMP against Geometric Fabrics, cuRobo, and OMPL RRTConnect. To provide
strong classical baselines, cuRobo and RRTConnect receive oracle scene geometry,
eight precomputed goal configurations, and relaxed mobile-base
motion limits. We use multiple goal configurations to avoid penalizing a planner
for a particular whole-body IK solution; for cuRobo, increasing from one to
eight precomputed goals raises success at a 12-s planning budget from 36.1\% to 45.6\%.

We select planning budgets using a paired sweep over 1,000 held-out queries.
With eight privileged goal configurations, cuRobo nearly saturates by 3~s
(44.6\% success), reaching only 46.0\% at 60~s. RRTConnect instead improves
from 9.2\% at 3~s to 81.0\% at 60~s. We therefore use 3~s for cuRobo and
60~s for RRTConnect, intentionally giving the latter substantially more
planning time (Appendix~\ref{app:classical_goal_time}). Geometric Fabrics also receives oracle geometry but plans
reactively. MoMa-NMP instead plans closed-loop from partial depth-camera and
LiDAR observations.

\subsection{Scene-Scale Whole-Body Planning}

\subsubsection{MoMa-NMP can solve globally constrained room-scale planning problems}

As shown in Table \ref{tab:mobile_mp_simulation}, MoMa-NMP succeeds on 87.0\% of held-out problems from partial sensor observations. RRTConnect reaches a comparable 85.3\%, but receives oracle scene geometry, eight precomputed goal configurations, relaxed base-motion limits, and up to 60~s to plan each query; successful plans require 12.5~s of planning time on average. In qualitative inspection, its trajectories are also typically longer and less smooth than those produced by MoMa-NMP. cuRobo and standalone Geometric Fabrics reach 46.9\% and 17.2\%, respectively. Thus, classical planning can achieve high success with privileged geometry and substantial computation, while MoMa-NMP amortizes this behavior into fast closed-loop planning from partial observations.

MoMa-NMP also retains performance as planning difficulty increases, reaching 84.4\% success on compartment-to-compartment problems and 85.1\% in scenes with at least four tables. Success decreases only from 90.3\% on one-table scenes to 84.2\% on seven-table scenes, compared with 73.2\% to 57.3\% for BC-only MoMa-NMP and 35.3\% to 6.6\% for Geometric Fabrics. These results show that MoMa-NMP retains global planning capability as the scene becomes more constrained.

\subsubsection{Comparison to \texorpdfstring{M$^2$Diffuser}{M2Diffuser}}

\begin{table}[t]
\caption{M$^2$Diffuser Benchmark Results}
\label{tab:m2diffuser}
\centering
\small
\setlength{\tabcolsep}{7pt}
\begin{tabular}{lcc}
\hline
Method &
\shortstack{Collision-free\\goal success (\%)} &
\shortstack{Goal reached\\(\%)} \\
\hline
M$^2$Diffuser & 23.2 & 27.8 \\
MoMa-NMP & \textbf{76.0} & \textbf{87.3} \\
\hline
\end{tabular}
\par\smallskip
\begin{minipage}{\linewidth}
\footnotesize
Paired evaluation on 891 endpoint-feasible problems from the official
M$^2$Diffuser seen-scene goal-reaching test split. MoMa-NMP is tested in a zero-shot manner without additional training.
\end{minipage}
\end{table}

We additionally evaluate MoMa-NMP on the official M$^2$Diffuser goal-reaching
benchmark without retraining or fine-tuning on M$^2$Diffuser scenes or demonstrations. We use the published M$^2$Diffuser evaluation criterion
\cite{yan2025m2diffuser}: goal reaching requires a final EEF pose within
$4\,\mathrm{cm}$ position error and $20^\circ$ orientation error. Because
MoMa-NMP uses a different robot embodiment, we restrict the paired comparison
to the 891 test problems whose mapped start and goal EEF poses admit
collision-free IK solutions for our robot. Both methods use the same scene geometries, start EEF poses, EEF goals, and benchmark evaluator. MoMa-NMP receives its usual simulated partial point clouds, while the released M$^2$Diffuser baseline receives an oracle point cloud.

As shown in Table~\ref{tab:m2diffuser}, MoMa-NMP reaches the goal without a prior environment or self-collision on 76.0\% of problems, compared with 23.2\% for
the released M$^2$Diffuser baseline. Even when collisions are ignored,
MoMa-NMP reaches the goal on 87.3\% of problems compared with 27.8\% for
M$^2$Diffuser. This indicates that the gap is not driven only by collision
avoidance: MoMa-NMP is substantially more successful at reaching the
commanded EEF goal on these whole-body planning problems.

\subsection{What Enables Scene-Scale Neural Motion Planning?}

We ablate the data-generation and learning choices used to transfer difficult DemoSmith trajectories into a closed-loop neural policy.

\subsubsection{Fabrics action annotation improves demonstration learnability}

We test DemoSmith's separation of global path discovery from imitation-target
generation by training directly on the original cuRobo actions instead of the
Fabrics annotations. As shown in Table~\ref{tab:demosmith_ablation}, BC success
falls from 64.8\% to 47.0\%, while world collision-before-success rises from
34.1\% to 53.0\%. Yet, when collisions are ignored, the policy trained on cuRobo actions still reaches the goal region in
99.7\% of scenes.

The difference also persists when each BC policy is used as the base policy for
the privileged Fabrics teacher used during fine-tuning. The teacher succeeds on
92.5\% of evaluation problems when guided by the Fabrics-trained BC policy,
compared with 88.0\% when guided by the raw-cuRobo BC policy. Thus, local Fabrics
correction does not eliminate the need for a strong learned global motion prior;
higher-quality Fabrics action targets improve both the behavior-cloned policy
and the corrective teacher built on top of it.

\begin{table}[t]
\caption{Action-Target Ablation}
\label{tab:demosmith_ablation}
\centering
\small
\setlength{\tabcolsep}{3.5pt}
\begin{tabular}{lccc}
\hline
Action targets &
\shortstack{BC success\\(\%)} &
\shortstack{BC collision\\(\%)} &
\shortstack{Teacher success\\(\%)} \\
\hline
Geometric Fabrics & \textbf{64.8} & \textbf{34.1} & \textbf{92.5} \\
cuRobo & 47.0 & 53.0 & 88.0 \\
\hline
\end{tabular}
\par\smallskip
\begin{minipage}{\linewidth}
\footnotesize
Both BC variants use the same trajectory corpus, architecture,
and observations (Appendix~\ref{app:action_target_ablation}). BC collision denotes world collision before success. Teacher success evaluates the privileged Fabrics teacher using the corresponding BC policy to provide its joint-space target sequence.
\end{minipage}
\end{table}

\subsubsection{On-policy refinement improves closed-loop robustness}

As shown in Table \ref{tab:mobile_mp_simulation}, privileged on-policy fine-tuning raises success from 64.8\% after behavior cloning to 87.0\% and reduces collision-before-success from 34.1\% to 12.3\%. On compartment-to-compartment tasks, success increases from 56.6\% to 84.4\%, showing that corrective supervision on policy-visited states is particularly important for long, constrained rollouts.

\subsubsection{Relative goal representations improve behavior cloning}

Removing $\Delta_{\mathrm{base}}$ and $\Delta_{\mathrm{eef}}$ reduces BC success from 64.8\% to 27.5\% and increases collision-before-success from 34.1\% to 56.9\% as seen in Table \ref{tab:training_ablations}.

\begin{table}[t]
\caption{Goal-Representation Ablation}
\label{tab:training_ablations}
\centering
\small
\setlength{\tabcolsep}{5pt}
\begin{tabular}{lcc}
\hline
Variant & Success (\%) & Collision (\%) \\
\hline
With delta-pose inputs & \textbf{64.8} & \textbf{34.1} \\
Without delta-pose inputs & 27.5 & 56.9 \\
\hline
\end{tabular}
\par\smallskip
\begin{minipage}{\linewidth}
\footnotesize
The ablated policy removes $\Delta_{\mathrm{base}}$, $\Delta_{\mathrm{eef}}$, and their MLP encoders, but retains the current and goal EEF poses and the rest of the behavior cloning pipeline.
\end{minipage}
\end{table}

\subsection{Real-World Generalization}

We evaluate MoMa-NMP on 20 motion-planning tasks across 10 unseen real-world scenes, with four trials per task. A trial is successful if the EEF reaches within $1\,\mathrm{cm}$ and $15^{\circ}$ of the goal without an environment or self-collision.

\subsubsection{MoMa-NMP generalizes to unseen constrained environments}

As seen in Table \ref{tab:real_world_results}, MoMa-NMP succeeds on 62/80 real-world trials (77.5\%), compared with 23/80 (28.8\%) for cuRobo and 17/80 (21.3\%) for RRTConnect. MoMa-NMP outperforms cuRobo on 16 of 20 tasks and RRTConnect on 17 of 20, demonstrating broad transfer from procedural DemoSmith scenes to unseen real-world geometry.

In the physical experiments, cuRobo and RRTConnect construct collision geometry from the initially-observed point cloud and receive one target EEF pose, with a 60~s planning budget per trajectory. Their trajectories are tracked using low-level feedback control, but the planners do not replan from updated observations. Iterative replanning increased collisions when robot-near point filtering removed nearby obstacle geometry, and RRTConnect was too slow for frequent replanning. These limitations help explain their performance gap between the oracle-geometry simulation and real-world experiments.

MoMa-NMP succeeds on motions that exit constrained workspaces, navigate around out-of-distribution obstacles or through narrow passages, and enter new constrained workspaces. It consistently plans base motion that moves tangentially to or away from the goal in order to avoid partially observed obstructions to a naive linear path. All 18 unsuccessful trials were collisions. Qualitative failures include nearly fully occluded obstacles and unfamiliar structures such as chairs.

\begin{table}[t]
\caption{Real-World Results}
\label{tab:real_world_results}
\centering
\small
\setlength{\tabcolsep}{6pt}
\begin{tabular}{lccc}
\hline
Method & Successes & Trials & Success (\%) \\
\hline
MoMa-NMP & \textbf{62} & 80 & \textbf{77.5} \\
cuRobo & 23 & 80 & 28.8 \\
RRTConnect & 17 & 80 & 21.3 \\
\hline
\end{tabular}
\par\smallskip
\begin{minipage}{\linewidth}
\footnotesize
Results span 20 tasks in 10 unseen scenes with four trials per task.
MoMa-NMP replans closed-loop from onboard observations; cuRobo and
RRTConnect plan once from collision geometry constructed from the initial
observation.
\end{minipage}
\end{table}

\subsubsection{Sensor-realistic training improves real-world transfer}

To evaluate training under partial visibility, we compare MoMa-NMP with a policy trained from complete scene point clouds. The observation type differs during both BC and fine-tuning, and each fine-tuning teacher is guided by its corresponding frozen BC policy. Both policies receive the same real depth-camera and LiDAR observations at deployment. In a separate cohort of 11 tasks across six scenes, sensor-realistic training improves success from 54.5\% to 79.5\% across 44 trials per policy (Table~\ref{tab:pointcloud_ablation}).

\begin{table}[t]
\caption{Training-Observation Ablation}
\label{tab:pointcloud_ablation}
\centering
\small
\setlength{\tabcolsep}{5pt}
\begin{tabular}{lccc}
\hline
Training observations & Successes & Trials & Success (\%) \\
\hline
Rendered depth + LiDAR & \textbf{35} & 44 & \textbf{79.5} \\
Complete point sample & 24 & 44 & 54.5 \\
\hline
\end{tabular}
\par\smallskip
\begin{minipage}{\linewidth}
\footnotesize
Both policies receive identical real depth-camera and LiDAR observations at
deployment. Training observation type differs during both training
stages. The cohort is separate from the 80-trial evaluation.
\end{minipage}
\end{table}

The largest qualitative differences often occur when obstacles are mostly visible but partially occluded. The complete-geometry policy frequently moves into the unobserved portion of these obstacles, while MoMa-NMP more reliably maintains clearance. This shows that reproducing sensor occlusion during training materially improves transfer to partial real-world observations.

\section{Conclusion and Limitations}
We introduced MoMa-DemoSmith and MoMa-NMP for scene-scale neural motion planning. MoMa-NMP achieves 87.0\% success in simulation and 77.5\% across unseen real-world scenes from partial onboard observations, substantially outperforming classical baselines in deployment. These results show that neural motion planning can support robust, long-horizon whole-body motion in large, constrained environments.

Like many neural methods, MoMa-NMP struggles in scenes that diverge significantly from the training distribution. For example, it often collides with chairs, which are not well represented by either the pillars or tables in our training set. Further expanding the distribution of training scenes would likely improve policy performance in additional scenes. Although MoMa-NMP demonstrates an ability to plan around partially occluded obstacles, it often collides with nearby small obstacles that are fully occluded from both the LiDAR and depth cameras.

\section*{Acknowledgment}
The authors thank Peiqi Liu for his help in conducting experiments.

\IEEEtriggeratref{35}
\bibliographystyle{IEEEtran}
\bibliography{references}

\input{appendix}

\end{document}

%% file: appendix.tex
\clearpage

\appendices

\section{Implementation Details}
\label{app:implementation}

This section expands the implementation description in
Section~\ref{sec:method}. We first describe the Geometric Fabrics controller
shared by demonstration annotation, on-policy supervision, and the standalone
baseline, and then give the observation, architecture, and optimization details
for MoMa-NMP.

\subsection{Geometric Fabrics Formulation}
\label{app:fabrics_implementation}

\subsubsection{Configuration space and composition}

Our Geometric Fabrics implementation follows the natural-form construction of
\cite{vanwyk2022fabrics}. The fabric state contains the planar base, seven
Franka joints, and six joints of the camera arm,
\begin{equation}
\mathbf{q}=
\left[x_b, y_b, \theta_b,
\mathbf{q}_{\mathrm{arm}}^{\mathsf T},
\mathbf{q}_{\mathrm{cam}}^{\mathsf T}\right]^{\mathsf T}
\in\mathbb{R}^{16}.
\label{eq:fabric_configuration}
\end{equation}
The learned policy controls only the first 10 coordinates (the base and Franka
arm); the camera-arm coordinates are attracted to a fixed nominal posture.

Each fabric term is defined in a task space
$\mathbf{x}_i=\phi_i(\mathbf{q})$ with Jacobian
$\mathbf{J}_i=\partial\phi_i/\partial\mathbf{q}$ and returns a natural form
$(\mathbf{M}_i,\mathbf{f}_i)$,
\begin{equation}
\mathbf{M}_i\ddot{\mathbf{x}}_i+\mathbf{f}_i=\mathbf{0},
\label{eq:fabric_natural_form}
\end{equation}
where the positive semidefinite metric $\mathbf{M}_i$ specifies the directional
priority of the term. The implementation maintains geometric ($g$) and forcing
or potential ($p$) channels separately. Their pullbacks are
\begin{align}
\mathbf{M}
&=\sum_{i,c}\mathbf{J}_i^{\mathsf T}\mathbf{M}_{i,c}\mathbf{J}_i,
\qquad c\in\{g,p\},
\label{eq:fabric_metric_pullback}\\
\mathbf{f}_{c}
&=\sum_i\mathbf{J}_i^{\mathsf T}
\left(\mathbf{f}_{i,c}
+\mathbf{M}_{i,c}\dot{\mathbf{J}}_i\dot{\mathbf{q}}\right).
\label{eq:fabric_force_pullback}
\end{align}
For nonlinear kinematic task maps, we evaluate the curvature term using the
directional finite difference
$\dot{\mathbf{J}}_i\approx
[\mathbf{J}_i(\mathbf{q}+\epsilon_J\dot{\mathbf{q}})
-\mathbf{J}_i(\mathbf{q})]/\epsilon_J$ with
$\epsilon_J=10^{-5}$. It is exactly zero for the linear configuration-space and
joint-limit maps.

The combined geometric acceleration is
$\ddot{\mathbf{q}}_g=-\mathbf{M}^{-1}\mathbf{f}_g$. Each selected energy task
map is assigned a Euclidean kinetic energy. After pullback and summation, the
root energy is
$E=\tfrac{1}{2}\dot{\mathbf{q}}^{\mathsf T}
\mathbf{M}_E\dot{\mathbf{q}}$. The transit controller attaches energy only to
the identity map, giving $\mathbf{M}_E=\mathbf{I}$ and
$\mathbf{f}_E=\mathbf{0}$. The goal-lock controller additionally attaches an
energy term to the stacked palm-point map, with Jacobian $\mathbf{J}_e$:
\begin{align}
\mathbf{M}_E&=\mathbf{I}+\mathbf{J}_e^{\mathsf T}\mathbf{J}_e,\\
\mathbf{f}_E&=\mathbf{J}_e^{\mathsf T}\dot{\mathbf{J}}_e\dot{\mathbf{q}}.
\end{align}
In the general natural form used by the implementation, the energization
coefficient is
\begin{equation}
\alpha=-\frac{\dot{\mathbf{q}}^{\mathsf T}
(\mathbf{M}_E\ddot{\mathbf{q}}_g+\mathbf{f}_E)}
{\dot{\mathbf{q}}^{\mathsf T}\mathbf{M}_E\dot{\mathbf{q}}+10^{-6}},
\label{eq:fabric_energization}
\end{equation}
where $\mathbf{f}_E$ includes the pulled-back curvature contribution when an
energy task map is nonlinear. The root natural force then becomes
\begin{align}
\mathbf{f}
&=\mathbf{f}_g-\alpha\mathbf{M}\dot{\mathbf{q}}+\mathbf{f}_p
\nonumber\\
&\quad+\left(d+\mathbb{I}[E>E_{\max}]d_s\right)
\mathbf{M}\dot{\mathbf{q}},
\label{eq:fabric_root_force}
\end{align}
where $d$ is global configuration-space damping and $d_s$ is additional damping
above the energy cap $E_{\max}$. We solve
$\ddot{\mathbf{q}}=-\mathbf{M}^{-1}\mathbf{f}$, then uniformly rescale each
batch element so that all coordinates satisfy their acceleration and
timestep-adjusted jerk limits. Finally, a constant-acceleration step advances
the fabric state,
\begin{align}
\mathbf{q}_{t+1}
&=\mathbf{q}_{t}+\Delta t\dot{\mathbf{q}}_{t}
+\tfrac{1}{2}\Delta t^2\ddot{\mathbf{q}}_{t},\\
\dot{\mathbf{q}}_{t+1}
&=\dot{\mathbf{q}}_{t}+\Delta t\ddot{\mathbf{q}}_{t}.
\label{eq:fabric_integration}
\end{align}

\subsubsection{Target attraction and posture}

For a configuration-space target, $\phi_i$ selects the base $xy$, base yaw,
seven Franka joints, or six camera-arm joints. Write
$\mathbf{e}=\mathbf{x}-\mathbf{x}^{\star}$, $r=\|\mathbf{e}\|_2$, and define
$\operatorname{unit}(\mathbf{e})=\mathbf{e}/\max(r,10^{-12})$. All attractors
used here have constant isotropic priority, $\mathbf{M}_p=m\mathbf{I}$.
Their desired acceleration and natural force are
\begin{align}
\mathbf{a}_{\mathrm{att}}
&=-k\tanh(\beta r)\operatorname{unit}(\mathbf{e})
-b(r)\dot{\mathbf{x}}, \label{eq:fabric_attraction}\\
b(r)&=\frac{b_0}{2}\left[1-\tanh\!\left(\lambda(r-r_d)\right)\right],
\label{eq:fabric_local_damping}\\
\mathbf{f}_p&=-\mathbf{M}_p\mathbf{a}_{\mathrm{att}}.
\end{align}
The first term is the negative gradient of
$\Phi(r)=(k/\beta)\log\cosh(\beta r)$. It is approximately linear near the
target, with stiffness $k\beta$, and approaches magnitude $k$ far away.
Thus, a distant target does not generate an arbitrarily large acceleration.
The damping approaches $b_0$ near the target and decreases outside radius
$r_d$; $\lambda$ sets the sharpness of this transition. The metric weight $m$
controls the term's influence relative to competing objectives after
pullback, rather than a physical robot mass.

Table~\ref{tab:fabric_terms} gives the values in these equations. Base translation and EEF-point distances are measured in meters, while rotational coordinates are measured in radians.
The same values are used across scenes.

\begin{table*}[t]
\caption{Attractor Parameters for Equations~\eqref{eq:fabric_attraction}--\eqref{eq:fabric_local_damping}}
\label{tab:fabric_terms}
\centering
\footnotesize
\setlength{\tabcolsep}{7pt}
\begin{tabular}{lcccccc}
\hline
Task and phase & Priority $m$ & Pull $k$ & Sharpness $\beta$ &
Damping $b_0$ & Transition $\lambda$ & Radius $r_d$ \\
\hline
Base $xy$, transit & 5 & 20 & 40 & 40 & 25 & 0.2 \\
Base $xy$, goal lock & 3 & 20 & 40 & 40 & 25 & 0.2 \\
Base yaw, transit & 3 & 20 & 40 & 25 & 10 & 0.3 \\
Base yaw, goal lock & 3 & 20 & 40 & 800 & 10 & 0.3 \\
Franka posture, transit & 0.5 & 100 & 40 & 200 & 10 & 0.2 \\
Franka posture, goal lock & 0.1 & 100 & 40 & 200 & 10 & 0.2 \\
Camera-arm posture & 0.1 & 50 & 40 & 100 & 10 & 0.2 \\
EEF points, goal lock only & 8 & 25 & 40 & 100 & 10 & 0.2 \\
\hline
\end{tabular}
\end{table*}

During transit, the base and Franka attractors follow the supplied full
configuration target. At goal lock, an additional task stacks the positions
of the palm center and six points $0.25\,\mathrm{m}$ along its positive and
negative coordinate axes. If $\mathbf{r}_{\ell}$ is one of these fixed palm-frame points, its
target is
$\mathbf{p}_{\ell}^{\star}=\mathbf{t}^{\star}
+\mathbf{R}^{\star}\mathbf{r}_{\ell}$ for the commanded EEF pose
$(\mathbf{R}^{\star},\mathbf{t}^{\star})$. Applying
Equation~\eqref{eq:fabric_attraction} to the stacked points penalizes both
position and orientation error through ordinary point kinematics. Its
Jacobian includes the base and arm, allowing either to contribute to EEF
correction. Joint-target attraction remains active but with weaker Franka
priority; the large yaw damping suppresses base rotation near the goal.

A separate, weak geometric posture term uses the full configuration target:
\begin{align}
\mathbf{M}_g&=10^{-3}\mathbf{I},\\
\mathbf{a}_g
&=-100\|\dot{\mathbf{q}}\|_2^2\tanh(20r_q)
\operatorname{unit}(\mathbf{q}-\mathbf{q}^{\star}),
\label{eq:fabric_posture_geometry}
\end{align}
where $r_q=\|\mathbf{q}-\mathbf{q}^{\star}\|_2$ and
$\mathbf{f}_g=-\mathbf{M}_g\mathbf{a}_g$ for this term. The squared-speed
factor makes it shape the motion direction without initiating motion from
rest; the forcing attractors supply that motion.

\subsubsection{Joint-limit avoidance}

For joint $j$, we create lower- and upper-limit distance maps,
$s_j^-=q_j-q_j^{\min}$ and $s_j^+=q_j^{\max}-q_j$. A positive distance is
inside the allowable interval. For either map, the implementation uses
\begin{align}
\gamma(s,\dot{s})&=\mathbb{I}[\dot{s}<0\ \text{or}\ s<0],\\
m_{\mathrm{lim}}(s,\dot{s})
&=\gamma(s,\dot{s})
\frac{0.02}{\max(s-0.1,\,2\times10^{-4})^2},
\label{eq:fabric_joint_limit_metric}\\
a_{\mathrm{lim}}(s,\dot{s})
&=8-16\,\mathbb{I}[\dot{s}\leq0]\dot{s}.
\label{eq:fabric_joint_limit_accel}
\end{align}
The task metric is diagonal with entries $m_{\mathrm{lim}}$, and the natural
force is $-m_{\mathrm{lim}}a_{\mathrm{lim}}$ coordinatewise. The gate gives
priority to a joint approaching or exceeding a limit. The metric grows
rapidly near the boundary, while the positive distance acceleration pushes
back into the feasible interval. Damping acts only on motion toward the
limit. The denominator floor bounds the metric numerically; it is not an
additional joint-position constraint. All distances in this term use the
joint's native units.

\subsubsection{Robot--environment and self-collision repulsion}

The robot is modeled with spheres, capsules, and rounded boxes. Scene
obstacles are procedural cuboids whose poses and dimensions are known to the
teacher. Each robot primitive's radius is inflated by $7.5\,\mathrm{mm}$.
This privileged geometry is used only by Fabrics and is not supplied to the
neural policy.

For an active primitive--obstacle or selected primitive--primitive pair
$\ell$, let $d_\ell$ be signed surface clearance and
$\mathbf{n}_\ell$ a unit vector pointing from the first robot body toward
the other body. World avoidance engages at
$d_\ell\leq0.1\,\mathrm{m}$; self-avoidance uses
$d_\ell\leq0.2\,\mathrm{m}$. Define the clamped clearance and directional
weight
\begin{align}
\bar d_\ell&=\operatorname{clip}(d_\ell,0.001,1.0),\\
\chi_\ell&=
\begin{cases}
1, & d_\ell\leq0,\\
\tfrac12[1-\tanh(1000(v_\ell-0.002))], & d_\ell>0,
\end{cases}
\label{eq:fabric_collision_gate}
\end{align}
where $v_\ell=-\mathbf{n}_\ell^{\mathsf T}\dot{\mathbf{c}}_\ell$ is the
first body's velocity away from the contact direction and
$\dot{\mathbf{c}}_\ell$ is its interpolated contact-point velocity.
The gate maintains priority for approaching contact and relaxes it when the
body moves away; the metric remains active after penetration.

Contact responses are distributed to robot-frame position slots. At a
capsule contact fraction $t$, the endpoint weights are $1-t$ and $t$; a box
response is applied at its center. Let $w_{k\ell}$ be the weight for slot
$k$, and let $\sigma_{k\ell}=+1$ on the first body and $-1$ on the second.
Summing over engaged contacts gives
\begin{align}
\mathbf{u}_k
&=\sum_\ell w_{k\ell}\sigma_{k\ell}
\frac{10}{\bar d_\ell}\mathbf{n}_\ell,\\
\mathbf{B}_k
&=\sum_\ell w_{k\ell}^{\,2}
\frac{10\chi_\ell}{\bar d_\ell}
\mathbf{n}_\ell\mathbf{n}_\ell^{\mathsf T}.
\label{eq:fabric_collision_blocks}
\end{align}
For self-collision, the opposite signs push the two bodies apart. The
rank-one matrices prioritize acceleration normal to the nearby surface;
motion tangent to that surface is less restricted. Cross-slot metric blocks
are omitted in this implementation.

Let $\mathbf{B}$ be the block-diagonal matrix of the $\mathbf{B}_k$, and
$d_k$ the smallest queried clearance incident on slot $k$. The forcing
metric and desired acceleration for that slot are
\begin{align}
\mathbf{M}_{p,k}
&=\frac{0.013}{\max(d_k,0.01)^2}
\frac{\mathbf{B}_k}{\|\mathbf{B}\|_F+10^{-6}},
\label{eq:fabric_collision_metric}\\
\mathbf{a}_{p,k}&=-3\,\operatorname{unit}(\mathbf{u}_k),\qquad
\mathbf{f}_{p,k}=-\mathbf{M}_{p,k}\mathbf{a}_{p,k}.
\label{eq:fabric_collision_acceleration}
\end{align}
The sums in Equation~\eqref{eq:fabric_collision_blocks} combine responses
from multiple nearby surfaces, weighting closer contacts more strongly.
The Frobenius normalization controls the overall scale of the contact
metric, and the coefficient $0.013$ sets its priority relative to target
attraction. The inverse-square factor further increases priority as
clearance decreases, with saturation below $1\,\mathrm{cm}$, while
normalizing $\mathbf{u}_k$ fixes the desired push-off acceleration magnitude
at $3\,\mathrm{m/s^2}$. The separate geometric
collision term and local collision damper are disabled; damping is supplied
by the root controller and target attractors.

World distances are evaluated against cuboid surfaces at five samples along
a capsule centerline or 15 samples on a robot box (its corners, face centers,
and center). The analytic cuboid path avoids triangle-mesh queries while
preserving this sampling rule. Segment--segment self distances use closest points;
segment--box and box--box pairs use sampled closest-point approximations.
These responses are soft motion preferences combined through the root
metric, rather than hard collision constraints.

\subsubsection{Energy regulation and integration limits}

In Equation~\eqref{eq:fabric_root_force}, $d=10$ dissipates motion energy
throughout the rollout. When $E>20$, an additional damping coefficient
$d_s=200$ reduces speed. This is an energy-based speed regulator in the chosen
task coordinates, not a hard bound on each joint velocity.

Let $a_j^{\max}$ and $j_j^{\max}$ denote the configured acceleration and
jerk limits for coordinate $j$. Before each integration step, the effective
acceleration bound is
$\min(a_j^{\max},\tfrac{1}{2}\Delta t\,j_j^{\max})$.
The controller uniformly scales the acceleration vector to satisfy these
bounds, preserving its direction rather than clipping coordinates
independently.

\subsubsection{Use within the pipeline}

The same natural-form implementation serves three distinct roles, summarized
in Table~\ref{tab:fabric_roles}. All three use oracle scene cuboids and robot
collision primitives. In all cases the action supplied to a
configuration-space attractor is held fixed during each integration substep.
When goal lock is enabled, it latches once the palm is within
$0.10\,\mathrm{m}$ and $30^{\circ}$ of the EEF goal; the EEF-pose fabric then
drives the terminal motion while retaining the policy command as a posture
reference.

\begin{table*}[t]
\caption{Role-Specific Geometric Fabrics Execution}
\label{tab:fabric_roles}
\centering
\footnotesize
\setlength{\tabcolsep}{4pt}
\begin{tabular}{p{0.16\textwidth}p{0.40\textwidth}p{0.16\textwidth}p{0.20\textwidth}}
\hline
Role & Configuration-space target & Goal lock & Integration \\
\hline
DemoSmith annotation &
The next five cuRobo base-and-arm configurations, supplied sequentially. &
Disabled. &
$60\,\mathrm{Hz}$; one step per target. \\

On-policy teacher &
The five configurations predicted by the frozen BC policy, corrected
sequentially from the visited state. &
Enabled. &
$60\,\mathrm{Hz}$ control; two $1/120\,\mathrm{s}$ substeps. \\

Standalone baseline &
The demonstration goal configuration until goal lock; the terminal phase uses
the EEF pose goal. &
Enabled. &
$60\,\mathrm{Hz}$ control; two $1/120\,\mathrm{s}$ substeps. \\
\hline
\end{tabular}
\end{table*}

\subsection{MoMa-NMP Observations and Actions}
\label{app:policy_inputs}

All geometric observations are expressed in a base-attached frame at the
Franka mounting origin. Both sensor clouds are cropped to a
$3\,\mathrm{m}$ horizontal radius, and LiDAR returns within $2\,\mathrm{cm}$ of the robot
collision model are removed. We allocate equal weight to the two sensors and
randomly pack 4,096 points from each into an 8,192-point sensed-scene cloud.
We independently sample 4,096 points from the current robot surface and append
them, producing 12,288 input points. An aligned scalar feature is zero for
sensed-scene points and one for sampled robot points. During training, robot
surface samples receive isotropic zero-mean Gaussian noise with standard
deviation equal to $0.5\%$ of the maximum robot-point radius in that sample.

\subsubsection{Sensor rendering}

The simulated observations are obtained by rasterizing sampled surfaces,
including the robot, rather than by ray tracing meshes. The camera keeps
the smallest forward depth in each pixel, and
the LiDAR keeps the smallest radial distance in each angular bin. Thus,
robot geometry can occlude obstacles before any robot-point filtering.
Table~\ref{tab:sensor_rendering} specifies the rendering parameters.

\begin{table*}[t]
\caption{Sensor Rendering Parameters}
\label{tab:sensor_rendering}
\centering
\footnotesize
\setlength{\tabcolsep}{5pt}
\begin{tabular}{p{0.20\textwidth}p{0.35\textwidth}p{0.35\textwidth}}
\hline
Setting & Depth camera & 3D LiDAR \\
\hline
Raster dimensions &
$H\times W=212\times120$ pixels &
512 polar bins $\times$ 512 azimuth bins \\
Angular coverage &
Horizontal $85.2^\circ$; vertical $58^\circ$ &
Azimuth $[-180^\circ,180^\circ)$; polar angle $[0^\circ,90^\circ)$ from local $+z$ \\
Near/far clipping &
$0.3$--$3.0\,\mathrm{m}$ forward depth &
$0.1$--$30.0\,\mathrm{m}$ radial range \\
Occlusion processing &
Nearest-depth splats expanded by two pixels in each direction
($5\times5$ footprint) &
Nearest-range bins with foreground suppression over a two-bin neighborhood \\
Return noise &
Independent Gaussian XYZ noise, $\sigma=4\,\mathrm{mm}$ &
Independent Gaussian XYZ noise, $\sigma=1\,\mathrm{mm}$ \\
\hline
\end{tabular}
\end{table*}

LiDAR foreground suppression reduces returns behind nearby surfaces.
Each sensor cloud is randomly subsampled to its fixed input budget;
valid points are repeated when too few returns are available. Visible robot
returns in the depth cloud remain sensed points with feature zero; only
the separately appended robot-surface samples receive feature one.

\subsubsection{State encoding and action convention}

The state input comprises the seven Franka joint angles and four 9D pose
features: current EEF pose, goal EEF pose, the goal transform expressed in the
base-aligned frame, and the goal transform expressed in the current EEF frame.
Each pose consists of a 6D rotation representation~\cite{zhou2019continuity}
and 3D translation. Franka joints are normalized by their joint limits; pose
translations and actions use statistics computed from the training set, while
6D rotations are not standardized.

The policy predicts a five-step sequence of 10D relative configurations. Each
token is relative to the configuration at the current policy query, rather
than relative to the preceding predicted token. Base $xy$ offsets are rotated
into the current base-yaw frame before normalization. At decoding, these
offsets are rotated back to the world frame and added to the current
configuration.

\subsection{Policy Architecture}
\label{app:policy_architecture}

Our architecture is adapted from IMPACT~\cite{yang2025drp} and uses a
PointNet++~\cite{qi2017pointnet++} point-cloud encoder.
Table~\ref{tab:policy_architecture} gives the complete learned architecture.
The checkpoint used for the primary results contains 8,696,082 trainable
parameters. Each modality encoder emits 124 features, which are concatenated
with a learned 4D modality embedding to form a 128D Transformer token.

\begin{table}[t]
\caption{MoMa-NMP Architecture}
\label{tab:policy_architecture}
\centering
\footnotesize
\setlength{\tabcolsep}{4pt}
\begin{tabular}{lp{0.60\columnwidth}}
\hline
Component & Configuration \\
\hline
Point input & 12,288 points with XYZ and one binary feature \\
PointNet++ & 256 centers; radius $0.1\,\mathrm{m}$; 64 neighbors; local MLP
$[4,64,64,64]$ \\
Point projection & $64\!\rightarrow\!248\!\rightarrow\!124$, LayerNorm and
LeakyReLU \\
State inputs & Joint token: 7D; four pose tokens: 9D each \\
State MLPs & $d_{\mathrm{in}}\!\rightarrow\!128\!\rightarrow\!256
\!\rightarrow\!124$, ReLU \\
Encoder input & 256 point tokens $+$ 5 state tokens \\
Transformer encoder & 10 layers; width 128; 8 heads; FFN width 2,048 \\
Transformer decoder & 6 layers; width 128; 8 heads; FFN width 1,024 \\
Action queries & 5 learned 128D tokens \\
Output head & Shared linear $128\!\rightarrow\!10$ \\
Regularization & Dropout $0$ \\
Parameter count & 8,696,082 \\
\hline
\end{tabular}
\end{table}

\subsection{Training Details}
\label{app:training_details}

Both stages minimize elementwise mean-squared error over the five predicted
10D offsets and use AdamW~\cite{loshchilov2019adamw} with
$(\beta_1,\beta_2)=(0.9,0.999)$, weight decay
$0.01$, and gradient-norm clipping at $1.0$. Forward and loss computation use
bfloat16 autocasting; model parameters, gradients, and optimizer state remain
in float32. Table~\ref{tab:training_hyperparameters} reports the remaining
stage-specific settings.

\begin{table*}[t]
\caption{MoMa-NMP Training Hyperparameters}
\label{tab:training_hyperparameters}
\centering
\footnotesize
\setlength{\tabcolsep}{5pt}
\begin{tabular}{p{0.19\textwidth}p{0.35\textwidth}p{0.37\textwidth}}
\hline
Setting & Behavior cloning & Privileged on-policy refinement \\
\hline
Supervision &
Precomputed five-step Fabrics chunks from approximately 6.7 million DemoSmith
trajectories. Invalid padded targets are masked. &
Five-step chunks obtained by sequentially correcting the frozen BC policy's
predictions with Fabrics. Targets after an intervening collision or reset are
masked. \\

Sampling / rollout &
The first and last 20 trajectory states are sampled with twice the weight of
intermediate states. &
Continuous rollouts with initial demonstration phase sampled uniformly from
$[0,1]$; environments reset independently after collision, goal dwell, stall,
or timeout. \\

Batching &
512 samples per rank; one optimizer update per sampled batch. &
2,048 environments and a 2,048-sample batch per GPU. Eight RTX A6000 GPUs give
16,384 simultaneous environments. \\

Learning rate &
$5\times10^{-4}$; 300-update linear warmup followed by cosine decay over the
nominal $10^6$-update schedule. &
$5\times10^{-5}$; linear warmup over two rollout episodes, then held constant. \\

Rollout length &
Not applicable. &
3,000 control steps per episode. A five-step action query is made every five
environment steps, giving 600 queries and updates per episode. \\

Rollout controller &
Not applicable. &
Teacher-controlled for 10 episodes, linearly transitioned from teacher to
student control for 10 episodes, and student-controlled thereafter, following
an expert-to-learner transition~\cite{ross2010efficient}. The
teacher and student are initialized from the same BC checkpoint, and the
teacher remains frozen. \\

Checkpoint used &
Final BC checkpoint: 360,301 optimizer updates. &
After 23 episodes: 69,000 environment steps and 13,800 policy
queries/optimizer updates. \\
\hline
\end{tabular}
\end{table*}

\subsubsection{On-policy supervision and rollout}
\label{app:on_policy_algorithm}

The frozen BC policy $\pi_0$ supplies the motion intent and the trainable
student $\pi_\theta$ supplies the deployment policy. At a visited state
$s_t$, both receive the same partial observation $o_t$ and decode five
absolute configuration targets. The teacher corrects $\pi_0$'s targets
sequentially using privileged geometry. These are five targets from one
frozen-policy query, rather than five new frozen-policy queries.

Let $F(s,\mathbf{a})$ denote one $1/60\,\mathrm{s}$ Fabrics correction of
target $\mathbf{a}$, using two integration substeps. A temporary teacher
rollout starts from the actual visited state:
\begin{align}
\tilde{s}_0&=s_t,\qquad
(\mathbf{a}_0^1,\ldots,\mathbf{a}_0^5)=\pi_0(o_t),\\
\mathbf{y}^h&=F(\tilde{s}_{h-1},\mathbf{a}_0^h),\\
\tilde{s}_h&=\operatorname{Step}(\tilde{s}_{h-1},\mathbf{y}^h),
\qquad h=1,\ldots,5.
\label{eq:teacher_chunk_rollout}
\end{align}
Each correction is anchored at the temporary simulator configuration and
its carried velocity. The velocity scale is one regardless of whether the
preceding executed chunk used the teacher or student. After each simulated
step, velocity is recomputed as the applied configuration displacement
divided by the control period. The teacher does not start from rest at
every target.

The temporary rollout checks robot--environment collision after each step.
The colliding target and all later targets in that chunk are excluded from
supervision; targets crossing a pending environment reset are also
excluded. If $m_{b,h}\in\{0,1\}$ denotes validity for environment $b$ and
chunk position $h$, the objective is
\begin{equation}
\mathcal{L}=
\frac{\sum_{b,h}m_{b,h}
\|\hat{\mathbf{z}}_{\theta,b}^{h}-\mathbf{z}_{b}^{h}\|_2^2}
{10\max(1,\sum_{b,h}m_{b,h})},
\label{eq:on_policy_masked_loss}
\end{equation}
where $\mathbf{z}_{b}^{h}$ encodes $\mathbf{y}_{b}^{h}$ as a normalized
10D offset from the original query configuration, using the action
convention above. The denominator counts valid scalar action coordinates
across all GPUs. Labels and validity masks are detached from the gradient
computation.

Teacher lookahead generates supervision without changing the state from
which the actual rollout proceeds. A separate Bernoulli teacher/student
choice is drawn for each environment at each query and held fixed across
its five targets, using the transition schedule in
Table~\ref{tab:training_hyperparameters}. This selects a complete controller
chunk rather than interpolating teacher and student configurations.

\begin{table*}[t]
\caption{One Query of Privileged On-Policy Refinement}
\label{tab:on_policy_algorithm}
\centering
\footnotesize
\setlength{\tabcolsep}{5pt}
\begin{tabular}{cp{0.87\textwidth}}
\hline
Step & Operation \\
\hline
1 & Reset completed environments, then form the current partial observation.
Predict the frozen BC and student five-target chunks without gradients. \\
2 & Draw one teacher/student controller choice per environment using the
training schedule. Save the rollout state for teacher lookahead. \\
3 & Sequentially correct the five BC targets using
Equation~\eqref{eq:teacher_chunk_rollout}; record corrected configurations and
collision/reset validity masks. Restore the saved state. \\
4 & Execute the selected teacher or student chunk for five control steps.
Reset environments on collision, sustained goal reach, stall, or timeout.
After a reset, hold that environment at its reset configuration until the
next query so that remaining targets from the old scene are not executed. \\
5 & Using the observation and teacher labels saved before execution,
take one AdamW update on Equation~\eqref{eq:on_policy_masked_loss}.
Discard this query's training batch and continue from the resulting rollout state. \\
\hline
\end{tabular}
\end{table*}

Table~\ref{tab:on_policy_algorithm} gives the implemented order. The first synchronized reset samples a phase uniformly along each
demonstration; subsequent independent resets start at the demonstration
beginning. Environments persist across training-episode boundaries. During on-policy training, we sample demonstrations from compartment-to-compartment tasks. Goal-dwell resets require ten consecutive
steps within $5\,\mathrm{cm}$ and $15^\circ$; these are training reset
tolerances, distinct from the $1\,\mathrm{cm}$ evaluation tolerance.
A stall is detected when the EEF changes by less than $1\,\mathrm{cm}$ and
$5^\circ$ over a 50-step window.

\section{Simulation Evaluation Details}
\label{app:simulation_details}

We sample 10,000 held-out start--goal problems from
successfully generated MoMa-DemoSmith demonstrations,
using the same generation pipeline as for the training data. 
We exclude 63 problems whose initial robot configurations
collide with the environment under the evaluation collision
representation, leaving 9,937 evaluation problems.
A rollout succeeds when the EEF first reaches within $1\,\mathrm{cm}$ position error and $15^{\circ}$ orientation error of the commanded goal without a preceding robot--environment collision. Robot link--link self-collision is not scored in this benchmark. Collision-before-success counts world collisions before successful goal reach. Rollouts terminate after 3,000 control steps at $60\,\mathrm{Hz}$. Extending the limit to 5,000 steps produces no additional successes.

\section{MoMa-DemoSmith Analysis}
\label{app:demosmith_analysis}

\subsection{Scene and Endpoint Sampling}
\label{app:demosmith_sampling}

DemoSmith draws single-table geometries from a pre-generated asset library
and composes them into larger scenes. Asset families include shelves,
open boxes, cubbies, cages, microwaves, dishwashers, cabinets, and drawers,
with additional cuboid clutter. The table library stores both obstacle
geometry and free compartment volumes. Table~\ref{tab:demosmith_sampling}
summarizes the scene diversity and endpoint proposals. Numerical ranges denote uniform sampling.

\begin{table*}[t]
\caption{DemoSmith Arrangement and Endpoint Sampling}
\label{tab:demosmith_sampling}
\centering
\footnotesize
\setlength{\tabcolsep}{5pt}
\begin{tabular}{p{0.24\textwidth}p{0.66\textwidth}}
\hline
Quantity & Distribution or construction \\
\hline
Single-table geometry &
Footprint $[0.8,1.0]\times[1.2,1.7]\,\mathrm{m}$, with clutter and
compartment geometry drawn from the asset library. \\
Arrangement family &
Line with probability $0.33$; otherwise a circle, inward- or outward-facing
with equal probability. \\
Line layout &
One to three tables with consecutive spacing of $[1.3,1.6]\,\mathrm{m}$. \\
Circular layout &
Initially three to seven tables, with radius increasing with table count.
Random table removal opens gaps while preserving the selected start and
goal tables. \\
Layout variation &
Perturb table positions, heights, and orientations to avoid fixed,
perfectly regular layouts. \\
Additional pillar &
Independently present near each table with probability $0.33$.
Height $[0.15,1.5]\,\mathrm{m}$; width $[0.06,0.2]\,\mathrm{m}$;
formed from overlapping cuboids. \\
Endpoint source &
Compartment sampling probability $0.95$ for the start and $0.75$ for the goal;
otherwise sample nearby free space. \\
Compartment candidates &
Sample interior positions with an $8\,\mathrm{cm}$ wall inset and tool
approach axes from the downward hemisphere. \\
Free-space candidates &
Sample positions in a volume around the selected table; reject positions
inside obstacle cuboids. \\
Base endpoint region &
Sample base positions near the selected table, with yaw concentrated
around its orientation. \\
\hline
\end{tabular}
\end{table*}

Start and goal table indices are sampled from the arrangement. On an
outward-facing circle they are distinct and separated by at most
$135^\circ$ around the circle; line and inward-facing layouts allow the
same table. Free-space orientations are obtained from independent Euler
angles in $[-\pi,\pi]$. These endpoint proposal
distributions are subsequently filtered by collision-aware IK and planning,
so their probabilities do not imply the same proportions in the accepted
trajectory dataset.

\subsection{Planning and Fabrics Annotation}
\label{app:demosmith_annotation}

The initial collision-aware IK solve uses 40 seeds per pose; the final goal
solve uses 80, with a $5\,\mathrm{cm}$ position convergence tolerance in
both stages. Failed endpoint queries are rejected. Before global planning, we express the goal base yaw as the equivalent angle closest to the starting base yaw. The global planner uses a $4\,\mathrm{mm}$ collision-sphere
buffer, base translation limits of $1.5\,\mathrm{m/s}$, and a base yaw limit
of $5\,\mathrm{rad/s}$. 
We set cuRobo's configuration-space acceleration limits to
$15\,\mathrm{m/s^2}$ for translational coordinates and
$15\,\mathrm{rad/s^2}$ for rotational coordinates, with
corresponding jerk limits of $500\,\mathrm{m/s^3}$ and
$500\,\mathrm{rad/s^3}$. These permissive planning limits are distinct from the
Fabrics execution limits specified above. The planning call has a $12\,\mathrm{s}$ budget, and accepted paths are resampled at $40\,\mathrm{Hz}$. A failed plan or a path with fewer than two
states is rejected; successful generation continues until the requested
number of accepted trajectories is obtained.

For annotation, each cuRobo configuration $\mathbf{q}_i$ serves as an
anchor whose next five path configurations supply the targets. The fabric
starts at $\mathbf{q}_i$ with
velocity
\begin{equation}
\dot{\mathbf{q}}_i^{\,\mathrm{seed}}=
\begin{cases}
\mathbf{0}, & i=0,\\
0.2(\mathbf{q}_i-\mathbf{q}_{i-1})/\Delta t_F, & i>0,
\end{cases}
\end{equation}
where $\Delta t_F=1/60\,\mathrm{s}$ is the export control period. The factor
$0.2$ is used only for the first correction at each anchor; subsequent
corrections carry the velocity of the temporary rollout with unit scaling.
This seed uses the export period, rather than the $1/40\,\mathrm{s}$
spacing of the original cuRobo path.

The exporter advances five sequential Fabrics corrections at $60\,\mathrm{Hz}$,
with one integration step per target and no EEF goal lock. Each corrected
configuration is encoded relative to the original anchor, not the preceding
corrected target. Available targets remain valid only while the temporary
rollout stays free of robot--environment collision; a collision invalidates
that target and the remaining chunk. Targets extending beyond the
trajectory end are also masked out. The
corrected actions are stored alongside the original trajectory observations:
annotation changes the supervision at each anchor, rather than replacing
the anchor observations with a single continuously refined trajectory.
The training corpus contains 6,658,695 saved global-trajectory records.

\subsection{Action-Target Ablation Protocol}
\label{app:action_target_ablation}

For the behavior-cloning comparison in Table~\ref{tab:demosmith_ablation},
both variants use the same underlying cuRobo trajectory corpus, policy
architecture, observations, five-target action horizon, and principal
optimization settings. The raw-cuRobo variant accumulates the stored
increments into five targets at the original $40\,\mathrm{Hz}$ trajectory
spacing without resampling to $60\,\mathrm{Hz}$. The Fabrics variant uses five
sequential corrections at $60\,\mathrm{Hz}$, masking collision-invalid targets
and subsequent steps in the affected chunk. The raw variant does not use
this Fabrics collision-validity mask. Normalization statistics are computed
separately for the two target distributions. Both policies execute their
predicted targets at the same $60\,\mathrm{Hz}$ rate during evaluation.

The reported raw-cuRobo and Fabrics checkpoints were evaluated after
242,201 and 360,301 training updates, respectively. The raw-cuRobo
policy's average success had already plateaued by about 145k updates;
the Fabrics policy achieved approximately 63.8\% success at about 240k
updates, compared with 64.8\% at its final checkpoint. Thus, the large
performance difference is not primarily explained by the additional
Fabrics training updates. This ablation compares the two supervision
pipelines, including their target timing and masking conventions.

\subsection{Global-Planner Backend Selection}
\label{app:planner_backend}
\label{app:graph_planner_ablations}

Generating approximately 6.7 million DemoSmith trajectories requires a planner
that can both solve difficult queries and generate accepted trajectories at
high throughput. We therefore compare cuRobo~\cite{sundaralingam2023curobo}
against five sampling-based planners implemented in
OMPL~\cite{sucan2012ompl}: RRTConnect~\cite{kuffner2000rrtconnect},
AIT*~\cite{strub2020ait}, ABIT*~\cite{strub2020abit},
BIT*~\cite{gammell2015bit}, and Informed RRT*~\cite{gammell2014informedrrt}.
All backends receive a common set of 3,840 post-IK planning queries from 120
procedurally generated scenes.

Each backend is evaluated over five seeds on one NVIDIA RTX~4090 and an AMD Threadripper~3960X. Parallelism is calibrated separately for each backend under a common GPU-memory constraint. To measure performance under the actual data-generation workload, we report accepted trajectories per planner-active second, including planning, trajectory extraction, backend validation, and integration overhead while excluding scene generation, IK, downstream Fabrics annotation, and output writing. cuRobo and the OMPL planners use a 12~s planning timeout.

\begin{table}[t]
\caption{DemoSmith Planner Throughput}
\label{tab:planner_backend}
\centering
\small
\setlength{\tabcolsep}{4pt}
\begin{tabular}{lccc}
\hline
Planner & Processes & Success (\%) & Accepted/s \\
\hline
cuRobo & 8 & 28.4 & $\mathbf{3.056 \pm 0.089}$ \\
RRTConnect & 40 & 48.9 & $1.642 \pm 0.019$ \\
AIT* & 40 & 25.3 & $0.839 \pm 0.016$ \\
ABIT* & 40 & 26.8 & $0.755 \pm 0.018$ \\
BIT* & 40 & 24.7 & $0.698 \pm 0.015$ \\
Informed RRT* & 40 & 19.4 & $0.547 \pm 0.005$ \\
\hline
\end{tabular}
\par\smallskip
\begin{minipage}{\linewidth}
\footnotesize
Throughput is mean $\pm$ sample standard deviation over five seeds at
backend-specific calibrated parallelism. Accepted/s denotes validated
trajectories per planner-active second.
\end{minipage}
\end{table}

Although RRTConnect solves more individual queries, cuRobo generates accepted trajectories at substantially higher throughput under the calibrated parallel configurations. cuRobo produces 3.056 accepted trajectories/s, $1.86\times$ the throughput of RRTConnect, motivating its selection as the global-path generator for the DemoSmith dataset of approximately 6.7 million trajectories.

\subsection{DemoSmith Generation and Annotation Throughput}
\label{app:demosmith_generation_throughput}

We benchmark steady-state global-trajectory generation on four NVIDIA
RTX~A5000 GPUs, with eight workers (two per GPU), CUDA MPS, 40 allocated
Slurm CPU units, and 250~GB RAM. The timed workload includes scene and
start--goal sampling, initial cuRobo IK, RMP goal regularization, final IK,
global planning, and unsuccessful queries. Each worker first saves four warmup
paths; after a common barrier, each of three repeats generates 512 accepted
trajectories. Generation times of 17.35, 16.00, and 16.15 minutes yield
442.7, 479.9, and 475.5 accepted trajectories per allocated GPU-hour. Pooled
steady-state throughput is \textbf{465.4 accepted trajectories per allocated
GPU-hour} (1,536 trajectories over 3.3001 allocated GPU-hours).

The same 1,536 records were subsequently Fabrics-annotated and validated.
In separate warmed single-GPU annotation runs, 512-record processing-and-output
times were 55.20, 49.82, and 49.88 seconds (9.28, 10.28, and 10.26
records/s). Annotation timing starts after the first batch loads and includes
subsequent loading and writing. Valid target-slot rates were 98.1\%, 98.2\%,
and 98.2\%, including unavailable trajectory-end targets.

\subsection{Effect of Demonstration Dataset Scale}
\label{app:data_scale}

We evaluate the effect of demonstration scale by training
otherwise identical behavior-cloning policies on 1.88M and
6.66M DemoSmith trajectories. Both are evaluated on the
same 9,937 held-out problems using their converged checkpoints.
\begin{table}[H]
\caption{Dataset-Scale Ablation}
\label{tab:data_scale_ablation}
\centering
\scriptsize
\setlength{\tabcolsep}{2.8pt}
\begin{tabular}{lrrrr}
\hline
Training trajectories &
\shortstack{Success\\(\%)} &
\shortstack{Collision\\(\%)} &
\shortstack{Comp.\ $\to$ comp.\\(\%)} &
\shortstack{$\geq 4$ tables\\(\%)} \\
\hline
1.88M & 48.5 & 51.3 & 38.5 & 42.8 \\
6.66M & \textbf{64.8} & \textbf{34.1} & \textbf{56.6} & \textbf{60.4} \\
\hline
\end{tabular}
\end{table}

Increasing the dataset from 1.88M to approximately 7M trajectories raises overall success from 48.5\% to 64.8\% and reduces collision-before-success from 51.3\% to 34.1\%. The gains persist on compartment-to-compartment and higher-table-count problems, indicating that additional DemoSmith demonstrations continue to improve learning of the scene-scale planning problems considered in this work.

\section{Classical Planner Ablations}
\label{app:classical_planner_ablation}

The main experiments intentionally evaluate classical global planners under favorable conditions. Because performance can depend strongly on the supplied goal configurations, planning budget, and scene representation, we separately ablate these choices below.

\subsection{Goal Redundancy and Planning-Time Budget}
\label{app:classical_goal_time}

A single DemoSmith goal configuration may unnecessarily constrain a classical planner because many whole-body configurations can realize the same EEF goal. Likewise, a short planning timeout may underestimate the solve rate of sampling-based planners. We therefore jointly vary the number of available goal configurations and the planning-time budget.

We evaluate both planners on the same deterministic, stratified subset of 1,000 queries drawn from the 9,937 valid simulation problems. In the multi-goal condition, each planner receives eight precomputed whole-body goal configurations satisfying the target EEF pose. Each set contains the original DemoSmith goal configuration and seven additional collision-validated IK solutions. The original configuration is retained even when rejected by the planner's collision model (100 of 9,937 queries in the full benchmark); all eight configurations satisfy the $1\,\mathrm{cm}$ and $15^{\circ}$ EEF goal tolerances. Goal generation occurs outside the planning budget, intentionally favoring the classical planner. Both planners receive oracle scene geometry and use the same relaxed mobile-base motion limits used during DemoSmith trajectory generation.

\begin{table*}[!t]
\caption{Goal and Planning-Budget Ablations}
\label{tab:classical_goal_time}
\centering
\scriptsize
\setlength{\tabcolsep}{3.5pt}
\begin{tabular}{llccccc}
\hline
Planner &
Goal condition &
\shortstack{Budget\\(s)} &
\shortstack{Success\\(\%)} &
\shortstack{Comp.\ $\to$ comp.\\success (\%)} &
\shortstack{$\geq 4$ tables\\success (\%)} &
\shortstack{Mean successful\\planning time (s)} \\
\hline
cuRobo &
8 privileged goals & 1  & 41.4 & 40.5 & 38.3 & 0.32 \\
cuRobo &
8 privileged goals & 3  & 44.6 & 43.2 & 45.5 & 0.44 \\
cuRobo &
8 privileged goals & 12 & 45.6 & 45.3 & 45.1 & 0.48 \\
cuRobo &
8 privileged goals & 30 & 45.5 & 45.1 & 45.8 & 0.51 \\
cuRobo &
8 privileged goals & 60 & 46.0 & 45.3 & 44.8 & 0.48 \\
cuRobo &
1 DemoSmith goal & 12 & 36.1 & 34.4 & 35.4 & 0.28 \\
\hline
RRTConnect &
8 privileged goals & 1  & 0.0 & 0.0 & 0.0 & --- \\
RRTConnect &
8 privileged goals & 3  & 9.2 & 7.2 & 9.4 & 2.49 \\
RRTConnect &
8 privileged goals & 12 & 50.5 & 44.7 & 49.5 & 5.93 \\
RRTConnect &
8 privileged goals & 30 & 69.6 & 66.3 & 66.8 & 9.48 \\
RRTConnect &
8 privileged goals & 60 & 81.0 & 77.8 & 81.6 & 14.25 \\
RRTConnect &
1 DemoSmith goal & 12 & 51.1 & 47.4 & 53.1 & 5.71 \\
\hline
\end{tabular}
\par\smallskip
\begin{minipage}{\linewidth}
\footnotesize
Results use a paired 1,000-query subset. Eight-goal conditions receive
precomputed goal configurations, with goal generation excluded
from the planning budget.
\end{minipage}
\end{table*}

Providing multiple goals materially improves cuRobo, increasing 12~s success from 36.1\% to 45.6\%. Additional planning time provides little benefit once these goals are available: the 3~s condition reaches 44.6\% success, compared with 46.0\% at 60~s. We therefore use the eight-goal, 3~s cuRobo configuration for the full-population comparison.

RRTConnect shows the opposite dependence on planning time. Success increases from 50.5\% at 12~s to 69.6\% at 30~s and 81.0\% at 60~s. We therefore use the eight-goal, 60~s RRTConnect configuration in the main comparison, intentionally giving the planner a large computation budget and privileged goal information.

\subsection{EEF-Goal End-to-End Planning}
\label{app:classical_eef_goal}

The privileged multi-goal condition gives the classical planners information unavailable to MoMa-NMP: several precomputed whole-body configurations that already realize the target EEF pose. We therefore additionally evaluate an end-to-end interface in which each planner receives only the target EEF pose and must generate candidate goal configurations online. IK generation and motion planning share the same 12~s wall-clock budget.

\begin{table}[t]
\caption{Effect of the goal interface under a 12~s planning budget on the paired 1,000-query subset. Values report success rate (\%).}
\label{tab:classical_eef_goal}
\centering
\small
\setlength{\tabcolsep}{6pt}
\begin{tabular}{lcc}
\hline
Planner &
\shortstack{8 privileged\\joint goals} &
\shortstack{EEF goal +\\online IK} \\
\hline
cuRobo & 45.6 & 48.2 \\
RRTConnect & 50.5 & 34.3 \\
\hline
\end{tabular}
\end{table}

Online IK reduces RRTConnect success from 50.5\% to 34.3\%,
while cuRobo success increases from 45.6\% to 48.2\%.
The effect of online goal generation therefore depends on
the planner.

\subsection{Classical Planning from Sensor-Derived Scene Geometry}
\label{app:classical_sensor_geometry}

The main classical baselines receive oracle scene geometry, but this information is unavailable during deployment. To test whether these planners can instead operate from a partial sensor observation, we construct an approximate collision world from a single simulated point-cloud snapshot. Observed points are voxelized at $5\,\mathrm{cm}$ resolution and converted into merged cuboid collision geometry. Final trajectories are validated against the held-out oracle geometry.

\begin{table}[t]
\caption{Scene-Representation Ablation}
\label{tab:classical_sensor_geometry}
\centering
\small
\setlength{\tabcolsep}{7pt}
\begin{tabular}{lcc}
\hline
Method & Oracle geometry (\%) & Sensor geometry (\%) \\
\hline
cuRobo & 36.2 & 6.5 \\
RRTConnect & 54.1 & 8.6 \\
\hline
\end{tabular}
\par\smallskip
\begin{minipage}{\linewidth}
\footnotesize
Sensor-derived collision geometry is constructed from one partial point-cloud
observation using $5\,\mathrm{cm}$ voxelization and merged cuboids. Returned
trajectories are validated against oracle geometry.
\end{minipage}
\end{table}

Replacing oracle geometry with the sensor-derived representation reduces cuRobo success from 36.2\% to 6.5\% and RRTConnect success from 54.1\% to 8.6\%. On compartment-to-compartment problems, sensor-derived success falls to 1.8\% for cuRobo and 2.2\% for RRTConnect. The corresponding collision-before-success rates are 6.2\% and 9.1\%.

\section{Additional Real-World Evaluation Details}
\label{app:real_world_details}

\subsection{Hardware Evaluation Protocol}
\label{app:hardware_protocol}

Hardware success uses the same $1\,\mathrm{cm}$ position and $15^{\circ}$ orientation tolerances as simulation. We impose no fixed rollout-duration limit. A trial fails upon robot--environment collision, self-collision, or an unrecoverable execution failure such as stalling or reaching a joint limit. Collisions are determined by human inspection; contact involving non-rigid dangling cables is excluded.

\subsection{Classical Baseline Deployment}
\label{app:hardware_classical}

For cuRobo and RRTConnect, we construct collision geometry from the onboard point-cloud observation at the beginning of each trial. Each planner receives one target EEF pose, without privileged whole-body goal configurations, and has a 60~s planning budget per trajectory. The resulting trajectory is tracked using low-level feedback control, but the planner does not replan from updated observations during execution. Initial conditions are kept nearly identical across methods. The initial collision representation omits occluded portions of obstacles, allowing a planned trajectory to intersect previously unseen geometry.

We also tested iterative replanning from updated observations. In these trials, replanning increased collisions because points near the robot must be removed to prevent robot geometry from entering the obstacle representation; this filtering can also remove nearby obstacle geometry, creating additional gaps in the reconstructed collision world. RRTConnect was additionally impractical for frequent replanning in these environments, often requiring more than 30~s to return a trajectory.

%% file: references.bib
@article{kavraki1996prm,
  title={Probabilistic Roadmaps for Path Planning in High-Dimensional Configuration Spaces},
  author={Kavraki, Lydia E. and {\v{S}}vestka, Petr and Latombe, Jean-Claude and Overmars, Mark H.},
  journal={IEEE Transactions on Robotics and Automation},
  volume={12},
  number={4},
  pages={566--580},
  year={1996},
  doi={10.1109/70.508439}
}

@inproceedings{kuffner2000rrtconnect,
  title={{RRT-Connect}: An Efficient Approach to Single-Query Path Planning},
  author={Kuffner, Jr., James J. and LaValle, Steven M.},
  booktitle={Proceedings of the IEEE International Conference on Robotics and Automation (ICRA)},
  volume={2},
  pages={995--1001},
  year={2000},
  doi={10.1109/ROBOT.2000.844730}
}

@inproceedings{ratliff2009chomp,
  title={{CHOMP}: Gradient Optimization Techniques for Efficient Motion Planning},
  author={Ratliff, Nathan and Zucker, Matt and Bagnell, J. Andrew and Srinivasa, Siddhartha},
  booktitle={Proceedings of the IEEE International Conference on Robotics and Automation (ICRA)},
  pages={489--494},
  year={2009},
  doi={10.1109/ROBOT.2009.5152817}
}

@inproceedings{kalakrishnan2011stomp,
  title={{STOMP}: Stochastic Trajectory Optimization for Motion Planning},
  author={Kalakrishnan, Mrinal and Chitta, Sachin and Theodorou, Evangelos and Pastor, Peter and Schaal, Stefan},
  booktitle={Proceedings of the IEEE International Conference on Robotics and Automation (ICRA)},
  pages={4569--4574},
  year={2011},
  doi={10.1109/ICRA.2011.5980280}
}

@article{schulman2014trajopt,
  title={Motion Planning with Sequential Convex Optimization and Convex Collision Checking},
  author={Schulman, John and Duan, Yan and Ho, Jonathan and Lee, Alex and Awwal, Ibrahim and Bradlow, Henry and Pan, Jia and Patil, Sachin and Goldberg, Ken and Abbeel, Pieter},
  journal={The International Journal of Robotics Research},
  volume={33},
  number={9},
  pages={1251--1270},
  year={2014},
  doi={10.1177/0278364914528132}
}

@article{mukadam2018gpmp2,
  title={Continuous-Time Gaussian Process Motion Planning via Probabilistic Inference},
  author={Mukadam, Mustafa and Dong, Jing and Yan, Xinyan and Dellaert, Frank and Boots, Byron},
  journal={The International Journal of Robotics Research},
  volume={37},
  number={11},
  pages={1319--1340},
  year={2018},
  doi={10.1177/0278364918790369}
}

@inproceedings{sundaralingam2023curobo,
  title={{CuRobo}: Parallelized Collision-Free Robot Motion Generation},
  author={Sundaralingam, Balakumar and Hari, Siva Kumar Sastry and Fishman, Adam and Garrett, Caelan and Van Wyk, Karl and Blukis, Valts and Millane, Alexander and Oleynikova, Helen and Handa, Ankur and Ramos, Fabio and Ratliff, Nathan and Fox, Dieter},
  booktitle={Proceedings of the IEEE International Conference on Robotics and Automation (ICRA)},
  pages={8112--8119},
  year={2023},
  doi={10.1109/ICRA48891.2023.10160765}
}

@article{cheng2021rmpflow,
  title={{RMPflow}: A Geometric Framework for Generation of Multitask Motion Policies},
  author={Cheng, Ching-An and Mukadam, Mustafa and Issac, Jan and Birchfield, Stan and Fox, Dieter and Boots, Byron and Ratliff, Nathan D.},
  journal={IEEE Transactions on Automation Science and Engineering},
  volume={18},
  number={3},
  pages={968--987},
  year={2021},
  doi={10.1109/TASE.2021.3053422}
}

@article{vanwyk2022fabrics,
  title={Geometric Fabrics: Generalizing Classical Mechanics to Capture the Physics of Behavior},
  author={Van Wyk, Karl and Xie, Mandy and Li, Anqi and Rana, Muhammad Asif and Babich, Buck and Peele, Bryan and Wan, Qian and Akinola, Iretiayo and Sundaralingam, Balakumar and Fox, Dieter and Boots, Byron and Ratliff, Nathan D.},
  journal={IEEE Robotics and Automation Letters},
  volume={7},
  number={2},
  pages={3202--3209},
  year={2022},
  doi={10.1109/LRA.2022.3143311}
}

@inproceedings{ichter2018sampling,
  title={Learning Sampling Distributions for Robot Motion Planning},
  author={Ichter, Brian and Harrison, James and Pavone, Marco},
  booktitle={Proceedings of the IEEE International Conference on Robotics and Automation (ICRA)},
  pages={7087--7094},
  year={2018},
  doi={10.1109/ICRA.2018.8460730}
}

@inproceedings{lai2021plannerflows,
  title={{PlannerFlows}: Learning Motion Samplers with Normalising Flows},
  author={Lai, Tin and Ramos, Fabio},
  booktitle={Proceedings of the IEEE/RSJ International Conference on Intelligent Robots and Systems (IROS)},
  pages={2542--2548},
  year={2021},
  doi={10.1109/IROS51168.2021.9636190}
}

@article{johnson2023vqmpt,
  title={Learning Sampling Dictionaries for Efficient and Generalizable Robot Motion Planning With Transformers},
  author={Johnson, Jacob J. and Qureshi, Ahmed H. and Yip, Michael C.},
  journal={IEEE Robotics and Automation Letters},
  volume={8},
  number={12},
  pages={7946--7953},
  year={2023},
  doi={10.1109/LRA.2023.3322087}
}

@inproceedings{qureshi2019mpnet,
  title={Motion Planning Networks},
  author={Qureshi, Ahmed H. and Simeonov, Anthony and Bency, Mayur J. and Yip, Michael C.},
  booktitle={Proceedings of the IEEE International Conference on Robotics and Automation (ICRA)},
  pages={2118--2124},
  year={2019},
  doi={10.1109/ICRA.2019.8793889}
}

@inproceedings{fishman2023mpinets,
  title={Motion Policy Networks},
  author={Fishman, Adam and Murali, Adithyavairavan and Eppner, Clemens and Peele, Bryan and Boots, Byron and Fox, Dieter},
  booktitle={Proceedings of the 6th Conference on Robot Learning},
  series={Proceedings of Machine Learning Research},
  volume={205},
  pages={967--977},
  year={2023},
  publisher={PMLR}
}

@inproceedings{huang2023scenediffuser,
  title={Diffusion-Based Generation, Optimization, and Planning in {3D} Scenes},
  author={Huang, Siyuan and Wang, Zan and Li, Puhao and Jia, Baoxiong and Liu, Tengyu and Zhu, Yixin and Liang, Wei and Zhu, Song-Chun},
  booktitle={Proceedings of the IEEE/CVF Conference on Computer Vision and Pattern Recognition (CVPR)},
  pages={16750--16761},
  year={2023},
  doi={10.1109/CVPR52729.2023.01607}
}

@inproceedings{carvalho2023motiondiffusion,
  title={Motion Planning Diffusion: Learning and Planning of Robot Motions with Diffusion Models},
  author={Carvalho, Jo{\~a}o and Le, An T. and Baierl, Mark and Koert, Dorothea and Peters, Jan},
  booktitle={Proceedings of the IEEE/RSJ International Conference on Intelligent Robots and Systems (IROS)},
  pages={1916--1923},
  year={2023},
  doi={10.1109/IROS55552.2023.10342382}
}

@inproceedings{dalal2025neuralmp,
  title={Neural {MP}: A Neural Motion Planner},
  author={Dalal, Murtaza and Yang, Jiahui and Mendonca, Russell and Khaky, Youssef and Salakhutdinov, Ruslan and Pathak, Deepak},
  booktitle={Proceedings of the IEEE/RSJ International Conference on Intelligent Robots and Systems (IROS)},
  pages={1--8},
  year={2025},
  doi={10.1109/IROS60139.2025.11246908}
}

@inproceedings{yang2025drp,
  title={Deep Reactive Policy: Learning Reactive Manipulator Motion Planning for Dynamic Environments},
  author={Yang, Jiahui and Liu, Jason Jingzhou and Li, Yulong and Khaky, Youssef and Shaw, Kenneth and Pathak, Deepak},
  booktitle={Proceedings of the 9th Conference on Robot Learning},
  series={Proceedings of Machine Learning Research},
  volume={305},
  pages={5056--5071},
  year={2025},
  publisher={PMLR}
}

@article{chamzas2022motionbenchmaker,
  title={{MotionBenchMaker}: A Tool to Generate and Benchmark Motion Planning Datasets},
  author={Chamzas, Constantinos and Quintero-Pe{\~n}a, Carlos and Kingston, Zachary and Orthey, Andreas and Rakita, Daniel and Gleicher, Michael and Toussaint, Marc and Kavraki, Lydia E.},
  journal={IEEE Robotics and Automation Letters},
  volume={7},
  number={2},
  pages={882--889},
  year={2022},
  doi={10.1109/LRA.2021.3133603}
}

@article{lu2024nrp,
  title={Neural Randomized Planning for Whole Body Robot Motion},
  author={Lu, Yunfan and Ma, Yuchen and Hsu, David and Cai, Panpan},
  journal={arXiv preprint arXiv:2405.11317},
  year={2024}
}

@article{zhang2025m3bench,
  title={{M$^3$Bench}: Benchmarking Whole-Body Motion Generation for Mobile Manipulation in {3D} Scenes},
  author={Zhang, Zeyu and Yan, Sixu and Han, Muzhi and Wang, Zaijin and Wang, Xinggang and Zhu, Song-Chun and Liu, Hangxin},
  journal={IEEE Robotics and Automation Letters},
  volume={10},
  number={7},
  pages={7286--7293},
  year={2025},
  publisher={IEEE}
}

@article{yan2025m2diffuser,
  title={{M$^2$Diffuser}: Diffusion-based Trajectory Optimization for Mobile Manipulation in {3D} Scenes},
  author={Yan, Sixu and Zhang, Zeyu and Han, Muzhi and Wang, Zaijin and Xie, Qi and Li, Zhitian and Li, Zhehan and Liu, Hangxin and Wang, Xinggang and Zhu, Song-Chun},
  journal={IEEE Transactions on Pattern Analysis and Machine Intelligence},
  year={2025},
  publisher={IEEE},
  doi={10.1109/TPAMI.2025.3553454}
}

@inproceedings{jiao2021vkc,
  title={Efficient Task Planning for Mobile Manipulation: a Virtual Kinematic Chain Perspective},
  author={Jiao, Ziyuan and Zhang, Zeyu and Wang, Weiqi and Han, David and Zhu, Song-Chun and Zhu, Yixin and Liu, Hangxin},
  booktitle={Proceedings of the IEEE/RSJ International Conference on Intelligent Robots and Systems (IROS)},
  pages={8288--8294},
  year={2021}
}

@inproceedings{yang2025mobipi,
  title={Mobi-$\pi$: Mobilizing Your Robot Learning Policy},
  author={Yang, Jingyun and Huang, Isabella and Vu, Brandon and Bajracharya, Max and Antonova, Rika and Bohg, Jeannette},
  booktitle={Proceedings of the 9th Conference on Robot Learning},
  series={Proceedings of Machine Learning Research},
  volume={305},
  pages={3516--3536},
  year={2025},
  publisher={PMLR}
}

@inproceedings{sundaresan2026homer,
  title={{HoMeR}: Learning In-the-Wild Mobile Manipulation via Hybrid Imitation and Whole-Body Control},
  author={Sundaresan, Priya and Malhotra, Rhea and Miao, Phillip and Yang, Jingyun and Wu, Jimmy and Hu, Hengyuan and Antonova, Rika and Engelmann, Francis and Sadigh, Dorsa and Bohg, Jeannette},
  booktitle={Proceedings of the IEEE International Conference on Robotics and Automation (ICRA)},
  pages={21847--21854},
  year={2026},
  doi={10.1109/ICRA57385.2026.11696261}
}

@inproceedings{jauhri2024actpermoma,
  title={Active-Perceptive Motion Generation for Mobile Manipulation},
  author={Jauhri, Snehal and Lueth, Sophie and Chalvatzaki, Georgia},
  booktitle={Proceedings of the IEEE International Conference on Robotics and Automation (ICRA)},
  pages={1413--1419},
  year={2024},
  doi={10.1109/ICRA57147.2024.10610714}
}

@article{ratliff2018riemannian,
  title={Riemannian motion policies},
  author={Ratliff, Nathan D and Issac, Jan and Kappler, Daniel and Birchfield, Stan and Fox, Dieter},
  journal={arXiv preprint arXiv:1801.02854},
  year={2018}
}

@inproceedings{qi2017pointnet++,
  title={{PointNet++}: Deep Hierarchical Feature Learning on Point Sets in a Metric Space},
  author={Qi, Charles Ruizhongtai and Yi, Li and Su, Hao and Guibas, Leonidas J},
  booktitle={Advances in Neural Information Processing Systems},
  volume={30},
  year={2017}
}

@inproceedings{ross2010efficient,
  title={Efficient reductions for imitation learning},
  author={Ross, St{\'e}phane and Bagnell, Drew},
  booktitle={Proceedings of the thirteenth international conference on artificial intelligence and statistics},
  pages={661--668},
  year={2010},
  organization={JMLR Workshop and Conference Proceedings}
}

@inproceedings{qureshi2018deepsmp,
  title={Deeply informed neural sampling for robot motion planning},
  author={Qureshi, Ahmed H and Yip, Michael C},
  booktitle={2018 IEEE/RSJ International Conference on Intelligent Robots and Systems (IROS)},
  pages={6582--6588},
  year={2018},
  organization={IEEE}
}

@inproceedings{molina2020learnlink,
  title={Learn and link: Learning critical regions for efficient planning},
  author={Molina, Daniel and Kumar, Kislay and Srivastava, Siddharth},
  booktitle={2020 IEEE International Conference on Robotics and Automation (ICRA)},
  pages={10605--10611},
  year={2020},
  organization={IEEE}
}

@inproceedings{ichter2020criticalprm,
  title={Learned critical probabilistic roadmaps for robotic motion planning},
  author={Ichter, Brian and Schmerling, Edward and Lee, Tsang-Wei Edward and Faust, Aleksandra},
  booktitle={2020 IEEE International Conference on Robotics and Automation (ICRA)},
  pages={9535--9541},
  year={2020},
  organization={IEEE}
}

@inproceedings{cheng2020sampling,
  title={Learning an optimal sampling distribution for efficient motion planning},
  author={Cheng, Richard and Shankar, Krishna and Burdick, Joel W},
  booktitle={2020 IEEE/RSJ International Conference on Intelligent Robots and Systems (IROS)},
  pages={7485--7492},
  year={2020},
  organization={IEEE}
}

@inproceedings{wu2024tidybot++,
  title={{TidyBot++}: An Open-Source Holonomic Mobile Manipulator for Robot Learning},
  author={Wu, Jimmy and Chong, William and Holmberg, Robert and Prasad, Aaditya and Gao, Yihuai and Khatib, Oussama and Song, Shuran and Rusinkiewicz, Szymon and Bohg, Jeannette},
  booktitle={Proceedings of the 8th Conference on Robot Learning},
  series={Proceedings of Machine Learning Research},
  volume={270},
  pages={3729--3741},
  year={2025},
  publisher={PMLR},
  url={https://proceedings.mlr.press/v270/wu25a.html}
}

@inproceedings{shaw2023leap,
  title={{LEAP Hand}: Low-Cost, Efficient, and Anthropomorphic Hand for Robot Learning},
  author={Shaw, Kenneth and Agarwal, Ananye and Pathak, Deepak},
  booktitle={Proceedings of Robotics: Science and Systems},
  year={2023},
  doi={10.15607/RSS.2023.XIX.089}
}

@inproceedings{zhou2019continuity,
  title={On the Continuity of Rotation Representations in Neural Networks},
  author={Zhou, Yi and Barnes, Connelly and Lu, Jingwan and Yang, Jimei and Li, Hao},
  booktitle={Proceedings of the IEEE/CVF Conference on Computer Vision and Pattern Recognition (CVPR)},
  pages={5745--5753},
  year={2019},
  doi={10.1109/CVPR.2019.00589}
}

@inproceedings{loshchilov2019adamw,
  title={Decoupled Weight Decay Regularization},
  author={Loshchilov, Ilya and Hutter, Frank},
  booktitle={Proceedings of the International Conference on Learning Representations (ICLR)},
  year={2019},
  url={https://openreview.net/forum?id=Bkg6RiCqY7}
}

@article{sucan2012ompl,
  title={The {Open Motion Planning Library}},
  author={{\c{S}}ucan, Ioan A. and Moll, Mark and Kavraki, Lydia E.},
  journal={IEEE Robotics \& Automation Magazine},
  volume={19},
  number={4},
  pages={72--82},
  year={2012},
  doi={10.1109/MRA.2012.2205651}
}

@inproceedings{strub2020ait,
  title={Adaptively Informed Trees ({AIT*}): Fast Asymptotically Optimal Path Planning through Adaptive Heuristics},
  author={Strub, Marlin P. and Gammell, Jonathan D.},
  booktitle={Proceedings of the IEEE International Conference on Robotics and Automation (ICRA)},
  pages={3191--3198},
  year={2020},
  doi={10.1109/ICRA40945.2020.9197338}
}

@inproceedings{strub2020abit,
  title={Advanced {BIT*} ({ABIT*}): Sampling-Based Planning with Advanced Graph-Search Techniques},
  author={Strub, Marlin P. and Gammell, Jonathan D.},
  booktitle={Proceedings of the IEEE International Conference on Robotics and Automation (ICRA)},
  pages={130--136},
  year={2020},
  doi={10.1109/ICRA40945.2020.9196580}
}

@inproceedings{gammell2015bit,
  title={Batch Informed Trees ({BIT*}): Sampling-Based Optimal Planning via the Heuristically Guided Search of Implicit Random Geometric Graphs},
  author={Gammell, Jonathan D. and Srinivasa, Siddhartha S. and Barfoot, Timothy D.},
  booktitle={Proceedings of the IEEE International Conference on Robotics and Automation (ICRA)},
  pages={3067--3074},
  year={2015},
  doi={10.1109/ICRA.2015.7139620}
}

@inproceedings{gammell2014informedrrt,
  title={Informed {RRT*}: Optimal Sampling-Based Path Planning Focused via Direct Sampling of an Admissible Ellipsoidal Heuristic},
  author={Gammell, Jonathan D. and Srinivasa, Siddhartha S. and Barfoot, Timothy D.},
  booktitle={Proceedings of the IEEE/RSJ International Conference on Intelligent Robots and Systems (IROS)},
  pages={2997--3004},
  year={2014},
  doi={10.1109/IROS.2014.6942976}
}

@inproceedings{niu2026scalable,
  title     = {Scalable Trajectory Generation for Whole-Body Mobile Manipulation},
  author    = {Niu, Yida and Chang, Xinhai and Liu, Xin and Jiao, Ziyuan and Zhu, Yixin},
  year      = {2026},
  booktitle = {Proceedings of the IEEE/CVF Conference on Computer Vision and Pattern Recognition (CVPR)},
  pages     = {1798--1808},
  url       = {https://yzhu.io/publication/tamp2026cvpr/paper.pdf}
}

@article{honerkamp2022learning,
  title   = {{N$^2$M$^2$}: Learning Navigation for Arbitrary Mobile Manipulation Motions in Unseen and Dynamic Environments},
  journal = {IEEE Transactions on Robotics},
  author  = {Daniel Honerkamp and Tim Welschehold and Abhinav Valada},
  volume  = {39},
  number  = {5},
  pages   = {3601--3619},
  year    = {2023},
  doi     = {10.1109/TRO.2023.3284346}
}

@article{994917fa14d74d25b22c7a0364270210,
  title   = {Local Reactive Control for Mobile Manipulators With Whole-Body Safety in Complex Environments},
  author  = {Chunxin Zheng and Yulin Li and Zhiyuan Song and Zhihai Bi and Jinni Zhou and Boyu Zhou and Jun Ma},
  year    = {2025},
  doi     = {10.1109/LRA.2025.3550733},
  volume  = {10},
  pages   = {4556--4563},
  journal = {IEEE Robotics and Automation Letters},
  number  = {5}
}

@inproceedings{d901a4cef1c6447c99e28ced87264572,
  title     = {Coupled Mobile Manipulation via Trajectory Optimization with Free Space Decomposition},
  author    = {Max Spahn and Bruno Brito and Javier Alonso-Mora},
  year      = {2021},
  doi       = {10.1109/ICRA48506.2021.9561821},
  pages     = {12759--12765},
  booktitle = {Proceedings of the 2021 IEEE International Conference on Robotics and Automation (ICRA)},
  publisher = {IEEE}
}
